# Variational Inference based on Robust Divergences

Futoshi Futami

The University of Tokyo, RIKEN
futami@ms.k.u-tokyo.ac.jp

Issei Sato

The University of Tokyo, RIKEN
sato@k.u-tokyo.ac.jp

Masashi Sugiyama

RIKEN, The University of Tokyo
sugi@k.u-tokyo.ac.jp

## Abstract

Robustness to outliers is a central issue in real-world machine learning applications. While replacing a model to a heavy-tailed one (e.g., from Gaussian to Student-t) is a standard approach for robustification, it can only be applied to simple models. In this paper, based on Zellner's optimization and variational formulation of Bayesian inference, we propose an outlier-robust pseudo-Bayesian variational method by replacing the Kullback-Leibler divergence used for data fitting to a robust divergence such as the $\beta$- and $\gamma$-divergences. An advantage of our approach is that superior but complex models such as deep networks can also be handled. We theoretically prove that, for deep networks with ReLU activation functions, the *influence function* in our proposed method is bounded, while it is unbounded in the ordinary variational inference. This implies that our proposed method is robust to both of input and output outliers, while the ordinary variational method is not. We experimentally demonstrate that our robust variational method outperforms ordinary variational inference in regression and classification with deep networks.

## 1 Introduction

Robustness is a fundamental topic in machine learning and statistics. Although specific definitions of robustness may be problem-dependent, a commonly shared notion is *"an insensitivity to small deviations from the assumptions"*, according to the seminal book by Huber and Ronchetti [2011]. Robustness to outliers is getting more important these days since recent advances in sensor technology give a vast amount of data with spiky noise and crowd-annotated data is full of human errors (Raykar et al. [2010], Zhang et al. [2016], Liu et al. [2012], Bonald and Combes [2017] ).

A standard approach to robust machine learning is a *model-based* method, which uses a heavier-tailed distribution such as the Student-t distribution instead of the Gaussian distribution as a likelihood function (Murphy [2012]). However, as pointed out in Wang et al. [2017], the model-based method is applicable only to a simple modeling setup.

To handle more complex models, we employ the optimization and variational formulation of Bayesian inference by Zellner [1988]. In this formulation, the posterior model is optimized to fit data under the Kullback-Leibler (KL) divergence, while it is regularized to be close to the prior. In this paper, we propose replacing the KL divergence for data fitting to a robust divergence, such as the $\beta$-divergence (Basu et al. [1998]) and the $\gamma$-divergence (Fujisawa and Eguchi [2008]).

Another robust Bayesian inference method proposed by Ghosh and Basu [2016] follows a similar line to our method, which adopts the $\beta$-divergence for pseudo-Bayesian inference. They rigorously analyzed the statistical efficiency and robustness of the method, and numerically illustrated its behavior for the Gaussian distribution.

Our work can be regarded as an extension of their work to variational inference so that more complex models such as deep networks can be handled. For deep networks with ReLU activation functions, we prove that the *influence function* (IF) (Huber and Ronchetti [2011]) of our proposed inference method is bounded, while it is unbounded in the ordinary variational inference. This implies that our proposed method is robust to both input and output outliers, while the ordinary variational method is not.

In Wang et al. [2017], another robust Bayesian inference method based on a *weighted likelihood* was proposed, where weights are drawn from their prior distribution. They also conducted IF analysis and showed that IF is bounded *asymptotically*. On the other hand, our method is guaranteed to have a bounded IF for finite samples. In addition, by using IF, we numerically show that influence to the predictive distribution by outliers is also bounded in our proposed method.

Finally, we experimentally demonstrate that our robust variational method outperforms ordinary variational inference in regression and classification with neural networks.

## 2 Preliminaries

In this section, we review preliminary materials on statistical inference.

## 2.1 Maximum Likelihood Estimation and Its Robust Variants

Let us consider the problem of estimating an unknown probability distribution[1] $p^*(x)$ from its independent samples $x_{1:N} = \{x_i\}_{i=1}^N$. To this end, we consider a parametric model $p(x;\theta)$ with parameter $\theta$, and minimize the generalization error measured by the KL divergence $D_{\mathrm{KL}}$ from $p^*(x)$ to $p(x;\theta)$:

$$D_{\mathrm{KL}}\left(p^*(x)\|p(x;\theta)\right) = \int p^*(x) \log\left(\frac{p^*(x)}{p(x;\theta)}\right) dx. \tag{1}$$

However, since $p^*(x)$ is unknown in practice, it is replaced with

$$D_{\mathrm{KL}}\left(\hat{p}(x)\|p(x;\theta)\right) = \mathrm{Const.} - \frac{1}{N}\sum_{i=1}^N \ln p(x_i;\theta), \tag{2}$$

where $\hat{p}(x) = \frac{1}{N}\sum_{i=1}^N \delta(x, x_i)$, is the empirical distribution and $\delta$ is the Dirac delta function. Minimizing this empirical Kullback-Leibler divergence is equivalent to *maximum likelihood estimation*. Equating the partial derivative of Eq.(2) with respect to $\theta$ to zero, we obtain the following estimating equation:

$$0 = \frac{1}{N}\sum_{i=1}^N \frac{\partial}{\partial \theta} \ln p(x_i;\theta). \tag{3}$$

Maximum likelihood estimation is sensitive to outliers because it treats all data points equally. To circumvent this problem, outlier-robust divergence estimation has been developed in statistics.

The *density power divergence*, which is also known as the $\beta$-divergence, is a vital example (Basu et al. [1998]). The $\beta$-divergence from functions $g$ to $f$ is defined as

$$D_\beta\left(g\|f\right) = \frac{1}{\beta}\int g(x)^{1+\beta}dx - \frac{\beta+1}{\beta}\int g(x)f(x)^\beta dx + \int f(x)^{1+\beta}dx. \tag{4}$$

The $\gamma$-divergence (Fujisawa and Eguchi [2008]) is another family of robust divergences:

$$D_\gamma\left(g\|f\right) = \frac{1}{\gamma(1+\gamma)} \ln \int g(x)^{1+\gamma}dx - \frac{1}{\gamma}\ln \int g(x)f(x)^\gamma dx + \frac{1}{1+\gamma}\ln \int f(x)^{1+\gamma}dx. \tag{5}$$

In the limit of $\beta \to 0$ and $\gamma \to 0$, both the $\beta$- and $\gamma$-divergences converge to the KL divergence:

$$\lim_{\beta \to 0} D_\beta\left(g\|f\right) = \lim_{\gamma \to 0} D_\gamma\left(g\|f\right) = D_{\mathrm{KL}}(g\|f). \tag{6}$$

---
[1] Although we focus on estimating density $p^*(x)$, almost the same discussion is possible for estimating conditional density $p^*(y|x)$, as explained in Section 3.

Similarly to maximum likelihood estimation, minimizing the $\beta$-divergence (or the $\gamma$-divergence) from empirical distribution $\hat{p}(x)$ to $p(x;\theta)$ gives an empirical estimator:

$$\arg\min_{\theta} D_\beta\left(\hat{p}(x) \| p(x;\theta)\right). \tag{7}$$

This yields the following estimating equation:

$$0 = \frac{1}{N}\sum_{i=1}^{N} p(x_i;\theta)^\beta \frac{\partial}{\partial \theta} \ln p(x_i;\theta) - \mathbb{E}_{p(x;\theta)}\left[p(x;\theta)^\beta \frac{\partial}{\partial \theta} \ln p(x_i;\theta)\right], \tag{8}$$

where the second term assures the unbiasedness of the estimator. The first term in Eq.(8) is the likelihood weighted according to the power of the probability density for each data point. Since the probability densities of outliers are usually much smaller than those of inliers, those weights effectively suppress the likelihood of outliers.

When $\beta = 0$, all weights become one and thus Eq.(8) is reduced to Eq.(3). Therefore, adjusting $\beta$ corresponds to controlling the trade-off between robustness and efficiency. See Appendices A and B for more details.

Eqs.(3) and (8) are called an M-estimator, and Eq.(8) is also called a Z-estimator (Huber and Ronchetti [2011], Basu et al. [1998], Van der Vaart [1998]). In various machine learning applications, those methods showed superior performance (Narayan et al. [2015], Samek et al. [2013], Cichocki et al. [2011]).

## 2.2 Bayesian Inference and Variational Methods

In Bayesian inference, parameter $\theta$ is regarded as a random variable, having prior distribution $p(\theta)$. With Bayes' theorem, the Bayesian posterior distribution $p(\theta|x_{1:N})$ can be obtained as

$$p(\theta|x_{1:N}) = \frac{p(x_{1:N}|\theta)p(\theta)}{p(x_{1:N})}. \tag{9}$$

Zellner [1988] showed that $p(\theta|x_{1:N})$ can also be obtained by solving the following optimization problem: [2]

$$\arg\min_{q(\theta)\in\mathcal{P}} L(q(\theta)), \tag{10}$$

where $\mathcal{P}$ is the set of all probability distributions, $-L(q(\theta))$ is the *evidence lower-bound* (ELBO),

$$L(q(\theta)) = D_{\text{KL}}(q(\theta)\|p(\theta)) - \int q(\theta)\left(-N d_{\text{KL}}\left(\hat{p}(x)\|p(x|\theta)\right)\right) d\theta,$$

---
[2] Zellner's formulation of Bayesian inference was also used for extending variational inference to constrained methods (Zhu et al. [2014], Koyejo and Ghosh [2013]).

and $d_{\mathrm{KL}}\left(\hat{p}(x)\|p(x|\theta)\right)$ denotes the *cross-entropy*:

$$d_{\mathrm{KL}}\left(\hat{p}(x)\|p(x|\theta)\right) = -\frac{1}{N}\sum_{i=1}^{N}\ln p(x_i|\theta). \tag{11}$$

Note that Bayes posterior (9) can be expressed as

$$p(\theta|x_{1:N}) = \frac{e^{-Nd_{\mathrm{KL}}(\hat{p}(x)\|p(x|\theta))}p(\theta)}{p(x_{1:N})}. \tag{12}$$

In practice, the optimization problem of Eq.(10) is often intractable analytically, and thus we need to use some approximation method. A popular approach is to restrict the domain of the optimization problem to a set of analytically tractable probability distributions $\mathcal{Q}$. Let us denote such a tractable distribution as $q(\theta;\lambda) \in \mathcal{Q}$, where $\lambda$ is a parameter. Then the optimization problem is expressed as

$$\underset{q(\theta;\lambda)\in\mathcal{Q}}{\arg\min}\, L(q(\theta;\lambda)). \tag{13}$$

This optimization problem is called *variational inference*(VI).

## 3 Robust Variational Inference based on Robust Divergences

In this section, we propose a robust variational inference method based on robust divergences.

As detailed in Appendix C, Eq.(10) can be equivalently expressed as

$$\underset{q(\theta)\in\mathcal{P}}{\arg\min}\, \mathbb{E}_{q(\theta)}[D_{\mathrm{KL}}\left(\hat{p}(x)\|p(x|\theta)\right)] + \frac{1}{N}D_{\mathrm{KL}}\left(q(\theta)\|p(\theta)\right). \tag{14}$$

The first term can be regarded as the expected likelihood (see Eq.(2)), while the second term "regularizes" $q(\theta)$ to be close to prior $p(\theta)$.

To enhance robustness to data outliers, let us replace the KL divergence in the expected likelihood term with the $\beta$-divergence:

$$\underset{q(\theta)\in\mathcal{P}}{\arg\min}\, \mathbb{E}_{q(\theta)}[D_{\beta}\left(\hat{p}(x)\|p(x|\theta)\right)] + \frac{1}{N}D_{\mathrm{KL}}\left(q(\theta)\|p(\theta)\right). \tag{15}$$

Note that Eq.(15) can be equivalently expressed as

$$\underset{q(\theta)\in\mathcal{P}}{\arg\min}\, L_{\beta}(q(\theta)), \tag{16}$$

where $-L_{\beta}(q(\theta))$ is the $\beta$-ELBO defined as

$$L_{\beta}(q(\theta) = D_{\mathrm{KL}}(q(\theta)\|p(\theta))$$
$$- \int q(\theta)\left(-Nd_{\beta}\left(\hat{p}(x)\|p(x|\theta)\right)\right)d\theta, \tag{17}$$

Table 1: Cross-entropies for robust variational inference.

| | Unsupervised | Supervised |
|---|---|---|
| $\beta$ | $-\frac{\beta+1}{\beta}\frac{1}{N}\sum_{i=1}^{N} p(x_i\|\theta)^\beta + \int p(x\|\theta)^{1+\beta}dx$ | $-\frac{\beta+1}{\beta}\left\{\frac{1}{N}\sum_{i=1}^{N} p(y_i\|x_i,\theta)^\beta\right\} + \left\{\frac{1}{N}\sum_{i=1}^{N}\int p(y\|x_i,\theta)^{1+\beta}dy\right\}$ |
| $\gamma$ | $-\frac{1}{N}\frac{\gamma+1}{\gamma}\sum_{i=1}^{N}\frac{p(x_i\|\theta)^\gamma}{\{\int p(x\|\theta)^{1+\gamma}dx\}^{\frac{\gamma}{1+\gamma}}}$ | $-\frac{1}{N}\frac{\gamma+1}{\gamma}\sum_{i=1}^{N}\frac{p(y_i\|x_i,\theta)^\gamma}{\{\int p(y\|x_i,\theta)^{1+\gamma}dy\}^{\frac{\gamma}{1+\gamma}}}$ |

and $d_\beta(\hat{p}(x)\|p(x|\theta))$ denotes the $\beta$-cross-entropy:

$$d_\beta(\hat{p}(x)\|p(x|\theta)) = -\frac{\beta+1}{\beta}\frac{1}{N}\sum_{i=1}^{N} p(x_i|\theta)^\beta + \int p(x|\theta)^{1+\beta}dx.$$

For its solution, we have the following theorem (its proof is available in Appendix D):

**Theorem 1** *The solution of Eq.(15) is given by*

$$q(\theta) = \frac{e^{-Nd_\beta(\hat{p}(x)\|p(x|\theta))}p(\theta)}{\int e^{-Nd_\beta(\hat{p}(x)\|p(x|\theta))}p(\theta)d\theta}. \tag{18}$$

Interestingly, the above expression of $q(\theta)$ is the same as the *pseudo posterior* proposed in Ghosh and Basu [2016]. Although the pseudo posterior is not equivalent to the *posterior distribution* derived by Bayes' theorem, the spirit of updating prior information by observed data is inherited (Ghosh and Basu [2016]). For this reason, we refer to Eq.(18) simply as a *posterior* in this paper. We discuss how prior information is updated in pseudo-Bayes-posteriors in Appendix E.

The optimization problem (15) is generally intractable. Following the same line as the discussion in Section 2.2, let us restrict the set of all probability distributions to a set of analytically tractable parametric distributions, $q(\theta;\lambda) \in \mathcal{Q}$. Then the optimization problem yields

$$\arg\min_{q(\theta;\lambda)\in\mathcal{Q}} L_\beta(q(\theta;\lambda)).$$

We call this method $\beta$-*variational inference* ($\beta$-VI).

We optimize objective function $L_\beta$ by black-box variational inference method and re-parameterization trick (Ranganath et al. [2014]). In our implementation, we estimate the gradient of the objective function (17) by Monte Carlo sampling.

So far, we focused on the unsupervised learning case and the $\beta$-divergence. Actually, we can easily generalize the above discussion to the supervised learning case and also to the $\gamma$-divergence, by simply replacing the cross-entropy with a corresponding one shown in Table 1. We denote the objective function for the $\gamma$-divergence as $L_\gamma$ in the same way as Eq.(17). Note that, there are several choices for the $\gamma$-cross-entropy, as detailed in Appendix H. Explicit expression of $L$, $L_\beta$, and $L_\gamma$ are summarized in Appendix F.

# 4 Influence Function Analysis

In this section, we analyze the robustness of our proposed method based on the *influence function* (IF) (Huber and Ronchetti [2011]). IFs have been used in robust statistics to study how much contamination affects estimated statistics.

## 4.1 Influence Function

First, we review the notion of IFs. Let $G$ be an empirical cumulative distribution of $\{x_i\}_{i=1}^n$:

$$G(x) = \frac{1}{n} \sum_{i=1}^n \Delta_{x_i}(x), \tag{19}$$

where $\Delta_x$ stands for the point-mass 1 at $x$. Let $G_{\varepsilon,z}$ be a contaminated version of $G$ at $z$:

$$G_{\varepsilon,z}(x) = (1-\varepsilon)G(x) + \varepsilon \Delta_z(x), \tag{20}$$

where $\varepsilon$ is a contamination proportion. For a statistic $T$ and empirical distribution $G$, IF at point $z$ is defined as follows (Huber and Ronchetti [2011]):

$$\text{IF}(z, T, G) = \left.\frac{\partial}{\partial \varepsilon} T(G_{\varepsilon,z}(x))\right|_{\varepsilon=0} = \lim_{\varepsilon \to 0} \frac{T(G_{\varepsilon,z}(x)) - T(G(x))}{\varepsilon}. \tag{21}$$

Intuitively, IF is a relative bias of a statistic caused by contamination at $z$.

## 4.2 Derivation of Influence Functions

Now we analyze how posterior distributions derived by VI are affected by contamination. In ordinary VI, we derive a posterior by minimizing Eq.(13). Let us consider an approximate posterior $q(\theta; m)$ parametrized by $m$. Then the objective function given by Eq.(13) can be regarded as a function of $m$ whose first-order optimality condition yields

$$0 = \left.\frac{\partial}{\partial m} L\right|_{m=m^*}. \tag{22}$$

For notational simplicity, we denote $q(\theta; m^*)$ by $q^*(\theta)$.

Referring to Eq.(21), $T$ corresponds to $m^*$, and $G$ is approximated empirically by the training dataset in VI. Then substituting Eq.(20) into Eq.(13) and using Eq.(21) and Eq.(22) yield the following theorem (its proof is available in Appendix F):

**Theorem 2** *When data contamination is given by Eq.(20), IF of ordinary VI is given by*

$$\left(\frac{\partial^2 L}{\partial m^2}\right)^{-1} \frac{\partial}{\partial m} \mathbb{E}_{q^*(\theta)} \left[D_{\text{KL}}(q^*(\theta)\|p(\theta)) + Nl(z)\right], \tag{23}$$

Table 2: Influence functions for robust variational inference.

| | Unsupervised | Supervised z=(x',y') |
|---|---|---|
| $l(z)$ | $\ln p(z\|\theta)$ | $\ln p(y'\|x',\theta)$ |
| $l_\beta(z)$ | $\frac{\beta+1}{\beta}p(z\|\theta)^\beta - \int p(x\|\theta)^{1+\beta}dx$ | $\frac{\beta+1}{\beta}p(y'\|x',\theta)^\beta - \int p(y\|x',\theta)^{1+\beta}dy$ |
| $l_\gamma(z)$ | $\frac{\gamma+1}{\gamma}\frac{p(z\|\theta)^\gamma}{\{\int p(x\|\theta)^{1+\gamma}dx\}^{\frac{\gamma}{1+\gamma}}}$ | $\frac{\gamma+1}{\gamma}\frac{p(y'\|x',\theta)^\gamma}{\{\int p(y\|x',\theta)^{1+\gamma}dy\}^{\frac{\gamma}{1+\gamma}}}$ |

IF of $\beta$-VI is given by

$$\left(\frac{\partial^2 L_\beta}{\partial m^2}\right)^{-1}\frac{\partial}{\partial m}\mathbb{E}_{q^*(\theta)}\left[D_{\mathrm{KL}}(q^*(\theta)\|p(\theta)) + Nl_\beta(z)\right], \tag{24}$$

and IF of $\gamma$-VI is given by

$$\left(\frac{\partial^2 L_\gamma}{\partial m^2}\right)^{-1}\frac{\partial}{\partial m}\mathbb{E}_{q^*(\theta)}\left[D_{\mathrm{KL}}(q^*(\theta)\|p(\theta)) + Nl_\gamma(z)\right], \tag{25}$$

where $l(z)$, $l_\beta(z)$, and $l_\gamma(z)$ are defined in Table 2.

Using these expressions, we analyze how estimated variational parameters can be perturbed by outliers. In practice, it is important to calculate $\sup_z |\mathrm{IF}(z,\theta,G)|$, because if it diverges, the model can be sensitive to small contamination of data.

### 4.3 Influence Function Analysis for Specific Models

In our analysis, we consider two types of outliers—outliers related to input $x$ and outliers related to output $y$. For true data generating distributions $p^*(x)$ and $p^*(y|x)$, input-related outlier $x_\mathrm{o}$ does not obey $p^*(x)$ and output-related outlier $y_\mathrm{o}$ does not obey $p^*(y|x)$. Below we investigate whether IFs are bounded even when $x_\mathrm{o} \to \infty$ or $y_\mathrm{o} \to \infty$.

Although general IF analysis has been extensively carried out in statistics (Huber and Ronchetti [2011]), few works exist focusing on specific models that we often use in recent machine learning applications. Based on this, we consider neural network models for regression and classification (logistic regression). In neural networks, there are parameters $\theta = \{W, b\}$ where outputs of hidden units are calculated by multiplying $W$ to input and then adding $b$. Our analysis shows that $\sup_z |\mathrm{IF}(z,b,G)|$ is always bounded (see Appendix I for details), and our exemplary analysis results for $\sup_z |\mathrm{IF}(z,W,G)|$ are summarized in Table 3.

From Table 3, we can confirm that ordinary VI is always non-robust to output-related outliers. As for input-related outliers, ordinary VI is robust for the "tanh"-activation function, but not for the ReLU and linear activation functions. On the other hand, IFs of our proposed method are bounded for all three activation functions including ReLU. We have further conducted IF analysis for the Student-t likelihood, which is summarized in Appendix I.

Table 3: Behavior of $\sup_z |\text{IF}(z, W, G)|$ in neural networks, "Regression" and "Classification" indicate the cases of ordinary VI, while "$\beta$- and $\gamma$-Regression" and "$\beta$- and $\gamma$-Classification" mean that we used $\beta$-VI or $\gamma$-VI. "Activation function" means the type of activation functions used. "Linear" means that there is no nonlinear transformation, inputs are just multiplied W and added b. $(x_\text{o} : U, y_\text{o} : U)$ means that IF is unbounded while $(x_\text{o} : B, y_\text{o} : U)$ means that IF is bounded for input related outliers, but unbounded for output related outliers.

| Activation function | Regression | $\beta$- and $\gamma$-Regression | Classification | $\beta$- and $\gamma$-Classification |
|---|---|---|---|---|
| Linear | $(x_\text{o} : U, y_\text{o} : U)$ | $(x_\text{o} : B, y_\text{o} : B)$ | $(x_\text{o} : U)$ | $(x_\text{o} : B)$ |
| ReLU | $(x_\text{o} : U, y_\text{o} : U)$ | $(x_\text{o} : B, y_\text{o} : B)$ | $(x_\text{o} : U)$ | $(x_\text{o} : B)$ |
| tanh | $(x_\text{o} : B, y_\text{o} : U)$ | $(x_\text{o} : B, y_\text{o} : B)$ | $(x_\text{o} : B)$ | $(x_\text{o} : B)$ |

Actually, in Bayesian inference, what we really want to know in the end is the *predictive distribution* at test point $x_\text{test}$:

$$p(x_\text{test}|x_{1:N}) = \int p(\theta|x_{1:N})p(x_\text{test}|\theta)d\theta$$
$$\approx \int q^*(\theta)p(x_\text{test}|\theta)d\theta.$$

Therefore, it is important to investigate how the predictive distribution is affected by outliers. If the training dataset is contaminated at a rate of $\epsilon$ at point $z$, we can analyze the effect of such data contamination on the predictive distribution by using IFs of the posterior distribution:

$$\frac{\partial}{\partial \epsilon}\mathbb{E}_{q^*(\theta)}\left[p(x_\text{test}|\theta)\right] = \frac{\partial \mathbb{E}_{q^*(\theta)}\left[p(x_\text{test}|\theta)\right]}{\partial m} \frac{\partial m^*(G_{\varepsilon,z}(x))}{\partial \varepsilon}, \qquad (26)$$

where $\frac{\partial m^*(G_{\varepsilon,z}(x))}{\partial \varepsilon}$ can be analyzed with the IFs derived above. Since analytical discussion on this expression is difficult, we numerically examined its behavior in Section 5.2.

The above expression looks similar to the ones derived in Giordano et al. [2015] and Koh and Liang [2017]. However, discussion in Giordano et al. [2015] focused on prior perturbation and expression in Koh and Liang [2017] is applicable only to maximum likelihood estimation. To our knowledge, ours is the first work to derive IFs of variational inference for data contamination.

## 5 Experiments

In this section, we report the experimental results of our proposed method on toy and benchmark datasets. In all the experiments, we used mean-field black-box VI combined with the Adam (Kingma and Ba [2014]) optimizer and assumed that the prior and approximated posterior are both Gaussian. Detailed experimental setups can be found in Appendix L.

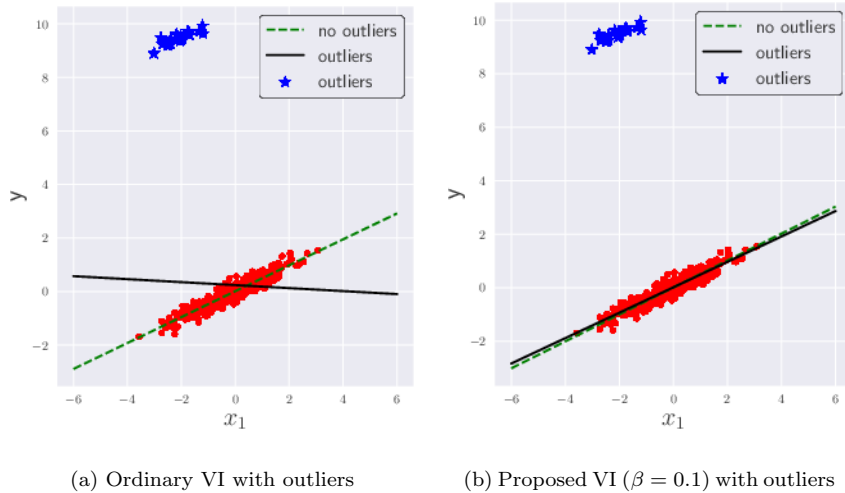

(a) Ordinary VI with outliers  (b) Proposed VI ($\beta = 0.1$) with outliers

Figure 1: Linear regression. Predictive distributions are derived by variational inference (VI).

## 5.1 Toy Data Experiment

We performed a toy dataset experiment for both regression and classification tasks to analyze the performance of the proposed method. We used a two-dimensional toy data and observed how the performance and the predictive distribution are affected by outliers when using ordinary VI and our method. The linear regression and logistic regression models are used. The detailed experimental setup is given in Appendix L.1.

For regression, the toy data and predictive distribution are shown in Fig. 1, where the horizontal axis indicates the first input feature $x_1$ and the vertical axis indicates the output $y$. As outliers, we considered input related outliers, which are caused by measurement error. The result of ordinary VI is heavily affected by outliers when there exist outliers, while the result of the proposed method is less affected by outliers.

For classification, we considered the situation where some of the labels are wrongly specified, as shown in Fig. 2. We also illustrated obtained decision boundaries in Fig. 2(a), which shows that the ordinary VI based method is heavily affected by outliers and Fig. 2(b) shows that our method with $\beta = 0.4$ is less affected by outliers.

## 5.2 Influence to Predictive Distribution

Based on Eq.(26), we numerically studied the influence of outliers on the predictive distribution. In this study, we used a two-hidden-layer neural network with 20 units in each hidden layer for regression and for classification with logistic loss.

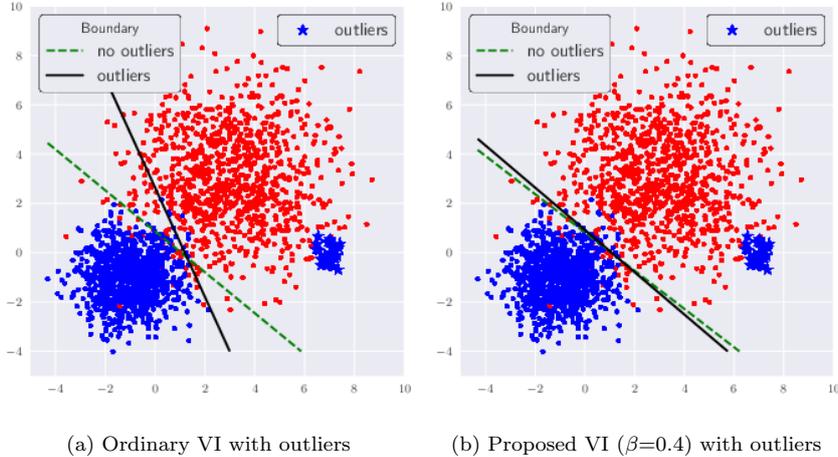

(a) Ordinary VI with outliers  (b) Proposed VI ($\beta$=0.4) with outliers

Figure 2: Boundaries of logistic regression using ordinary VI and the proposed method

**Regression**

We used the powerplant dataset in UCI (Lichman [2013]) which has four features for each input. Since it is difficult to visualize the behavior of the influence of predictive distributions, instead, we plot how the log-likelihood of a test point is influenced by an outlier. We compared the influence of ordinary VI based method and proposed method ($\beta$=0.1). To calculate Eq.(26), we have to specify an outlier and a test data point. As an input related outlier, we randomly chose a single data point from the training data and moved the first feature of the chosen data from $-\infty$ to $+\infty$. Similarly, as an output related outlier, we moved randomly chosen output $y$ from $-\infty$ to $+\infty$. As the test data point, we randomly chose a single data point from the test data. For the detailed experimental setting, see Appendix L.2.

The results are shown in Fig. 3, where the horizontal axis indicates the value of the perturbed feature, and the vertical axis indicates the value of $\frac{\partial}{\partial \epsilon} \mathbb{E}_{q^*(\theta)}[\ln p(x_{\text{test}}|\theta)]$.

The results in Fig. 3 show that the model using the ReLU activation inferred by ordinary VI can be affected infinitely by input related outliers, while the influence is bounded in our method. As for output related outliers, models inferred by ordinary VI are infinitely influenced, while influence in our method is bounded. From those results, we can see that our method is robust for both input and output related outliers in the sense that test point prediction is not perturbed infinitely by contaminating a single training point.

A notable difference from the IF analysis in Section. 4.3 is that for the perturbation by input related outliers for the tanh activation function, the value of of $\frac{\partial}{\partial \epsilon} \mathbb{E}_{q^*(\theta)}[\ln p(x_{\text{test}}|\theta)]$, does not converge to zero even for the proposed method in the limit that the absolute value of the input related outlier goes to $\infty$.

This might be due to the fact that in the limit, the input to the next layer goes to $\pm 1$ when the tanh activation function is used. For the next layer, an input which has value $\pm 1$ might not be so strange compared to regular data, and thus it is not regarded as an outlier. Therefore, during the optimization process, the likelihood of input related outliers is not downweighted so much in the robust divergence and the influence of outliers remains non-zero. If we use the ReLU activation function, in the limit, the input to the next layer becomes much larger than the regular data, and thus it is regarded as an outlier.

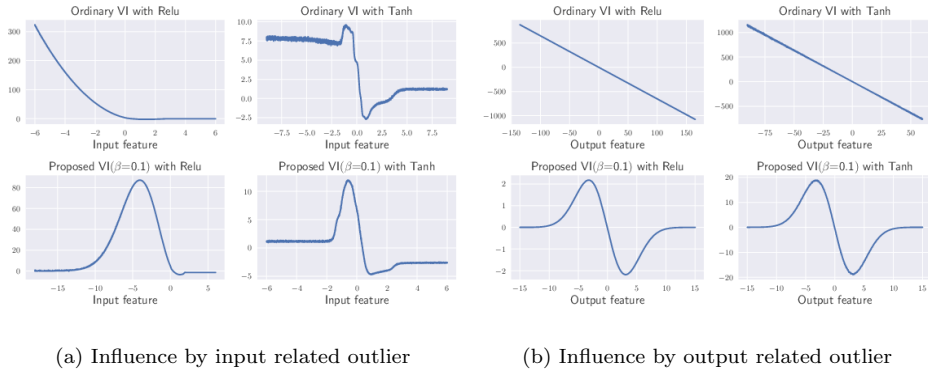

(a) Influence by input related outlier  (b) Influence by output related outlier

Figure 3: Influence on the test log-likelihood for neural net regression. The horizontal axis indicates the value of the perturbed feature, while the vertical axis indicates the value of $\frac{\partial}{\partial \epsilon}\mathbb{E}_{q^*(\theta)}[\ln p(x_{\text{test}}|\theta)]$.

**Classification**

We used the eeg dataset in UCI which has 14 features as input. In the same way as the regression experiment, as an input related outlier, we randomly chose a single data point from the training data and moved the third feature of the chosen data from $-\infty$ to $+\infty$. The result of how the test log-likelihood is influenced is given in Fig. 4. For ordinary VI, using the ReLU activation function causes unbounded influence, while our method keeps the influence bounded. We can also confirm that the influence in our method converges to smaller value than that in ordinary VI in the limit even in the case of tanh.

As an output related outlier, we investigated the influence of label misspecification. We flipped one of the labels in the training data and observed how the test log-likelihood changes. By assuming $\epsilon = \frac{1}{N}$, where $N$ is the number of training data, we calculated $\frac{1}{N}\frac{1}{N}\sum_i \frac{1}{N_{\text{test}}}\sum_j \frac{\partial}{\partial \epsilon_i}\mathbb{E}_{q^*(\theta)}\left[\ln p(y_{\text{test}}^j|x_{\text{test}}^j,\theta)\right]$, which represents the averaged amount of change in the test log-likelihood, and the term inside the sum over $j$ means the change in the log-likelihood for the $j$th test data caused by flipping the label of the $i$th training data. Without IF, this is difficult to calculate because we have to retrain a neural network with flipped data and this is extremely demanding .

Table 4 shows that the change in the test log-likelihood in our method is smaller than that in ordinary VI. This implies that our method is robust against label misspecification.

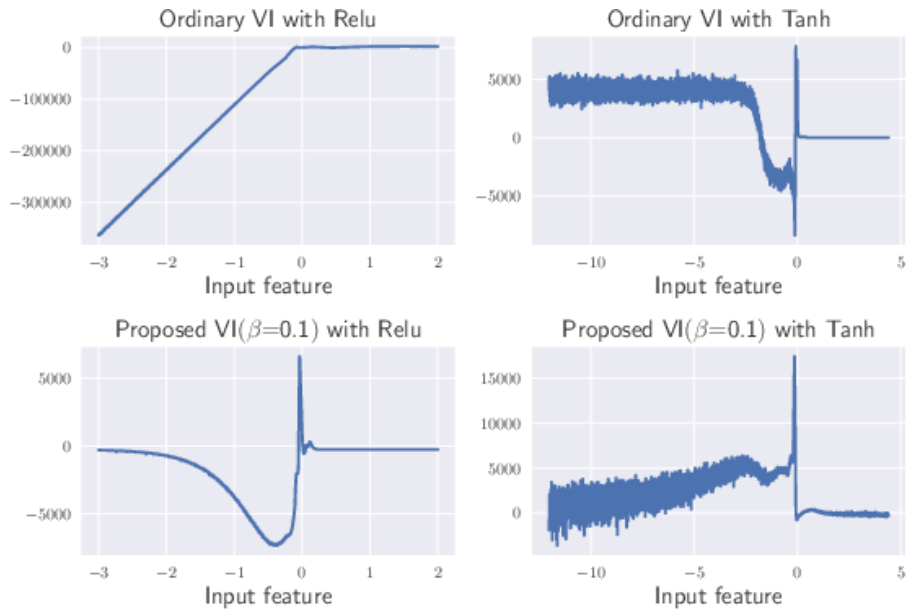

Figure 4: Influence on the test log-likelihood by input related outlier for neural net classification with logistic loss.

Table 4: Average change in the test log-likelihood

|      | Ordinary VI | Proposed VI ($\beta = 0.1$) |
| ---- | ----------- | --------------------------- |
| ReLU | -1.65e-3    | -3.29e-5                    |
| tanh | -2.3e-3     | -3.49e-4                    |

From these case studies, we confirmed that our method is robust for both input and output related outliers in both regression and classification settings in the sense that the prediction is less influenced by outliers.

## 5.3 How to Determine $\beta$ and $\gamma$

Finally we show that by choosing parameters $\beta$ and $\gamma$ by cross validation, our method can achieve even better performance compared to ordinary VI and other existing robust methods on several benchmark datasets in UCI. The detailed experimental setup is described in Appendix L.3.

**Regression**

We used a neural net which has two hidden layers each with 20 units and the ReLU activation function. As outliers, we added both input and output related outliers. The experimental results are summarized in Table 5. In Table 5, "Outliers" means the percentage of outliers in the training dataset we contaminated artificially. KL(G) means ordinary VI with the Gaussian likelihood, KL(St) is ordinary VI with the Student-t likelihood, WL means the method proposed in Wang et al. [2017], Rényi is the Rényi divergence minimization method proposed in Li and Turner [2016] and BB-$\alpha$ is the black-box $\alpha$ divergence minimization method proposed in Hernandez-Lobato et al. [2016] and Li and Gal [2017].

Table 5: Test regression accuracy in RMSE

| Dataset | Outliers | KL(G) | KL(St) | WL | Rényi | BB-$\alpha$ | $\beta$ | $\gamma$ |
|---|---|---|---|---|---|---|---|---|
| **concrete** | 0% | 8.87(2.57) | **7.34(0.41)** | 7.89(0.77) | 7.62(0.44) | **7.34(0.31)** | 7.58(0.38) | **7.34(0.76)** |
| $N$=1030 | 10% | 15.7(2.50) | 8.94(2.65) | 12.3(2.41) | 14.2(1.74) | 11.4(2.69) | **8.11(0.89)** | 8.26(0.98) |
| $D$=8 | 20% | 16.8(0.70) | 11.1(3.78) | 14.3(2.91) | 15.6(1.90) | 11.9(2.64) | **8.15(0.99)** | 9.25(1.27) |
| **powerplant** | 0% | 4.41(0.13) | 4.43(0.15) | 4.46(0.17) | 4.48(0.15) | 4.38(0.83) | **4.37(0.15)** | 4.45(0.17) |
| $N$=9568 | 10% | 6.44(1.88) | 4.54(0.14) | 5.12(0.41) | 5.49(0.45) | 5.91(1.63) | **4.39(0.14)** | 4.47(0.16) |
| $D$=4 | 20% | 9.97(4.7) | 4.56(1.45) | 6.44(0.52) | 6.87(1.09) | 5.52(1.31) | **4.41(0.15)** | 4.53(1.46) |
| **protein** | 0% | 5.61(0.38) | **4.79(0.05)** | 5.50(0.62) | 5.62(0.25) | 4.89(0.05) | 4.86(0.05) | **4.79(0.04)** |
| $N$=45730 | 10% | 6.13(0.02) | 4.92(0.05) | 6.13(0.03) | 6.11(0.03) | 6.13(0.03) | 4.91(0.04) | **4.90(0.06)** |
| $D$=9 | 20% | 6.14(0.03) | 4.98(0.07) | 6.14(0.03) | 6.12(0.03) | 6.10(0.28) | 4.96(0.05) | **4.95(0.06)** |

Table 6: Test classification accuracy

| Dataset | Outliers | KL | KL($\epsilon$) | WL | Rényi | BB-$\alpha$ | $\beta$ | $\gamma$ |
|---|---|---|---|---|---|---|---|---|
| **spam** | 0% | 90.9(5.8) | 91.2(4.4) | 89.2(5.7) | 90.0(0.7) | 92.9(1.5) | **93.3(1.3)** | 92.2(0.8) |
| $N$=4601 | 10% | 76.5(37.6) | 90.0(5.1) | 89.1(5.7) | **92.6(1.4)** | 91.6(1.4) | 92.4(1.2) | 92.1(1.1) |
| $D$=57 | 20% | 60.6(48.3) | 89.8(5.5) | 88.3(5.3) | 91.6(1.6) | 91.6(1.6) | **92.2(1.3)** | 91.6(1.4) |
| **eeg** | 0% | 72.8(2.9) | 77.7(3.2) | **81.3(2.4)** | 68.4(7.9) | 77.5(3.3) | 75.9(5.5) | 80.2(3.4) |
| $N$=14890 | 10% | 56.0(2.6) | 62.7(0.09) | 56.0(2.4) | 57.5(9.6) | 67.9(8.2) | 60.8(8.1) | **72.5(2.6)** |
| $D$=14 | 20% | 56.0(2.7) | 60.0(7.1) | 56.0(2.4) | 57.7(2.4) | 67.4(8.8) | 56.0(2.4) | **72.2(6.4)** |
| **covertype** | 0% | 65.2(8.8) | 73.1(6.2) | **73.4(6.3)** | 72.0(6.6) | 73.2(4.8) | 70.5(5.9) | **73.4(6.1)** |
| $N$=581012 | 10% | 60.2(16.9) | **74.4(6.2)** | 73.7(5.5) | 65.4(8.5) | 70.6(5.9) | 65.7(9.0) | 72.4(7.7) |
| $D$=54 | 20% | 56.4(18.7) | 71.4(10.4) | 71.2(7.2) | 67.6(9.7) | 67.1(8.1) | 66.2(9.6) | **72.3(5.9)** |

Our method compares favorably with ordinary VI and existing robust methods for all the datasets.

**Classification**

We used a neural net which has two hidden layers each with 20 units except for the covertype dataset. For the covertype dataset, we used a neural net which has one hidden layer with 50 units. We used the ReLU activation function for all the networks. As outliers, we considered both input and output related outliers. The experimental results are in Table 6. In Table 6, KL($\epsilon$) means that we used the robust loss function which is $p(y=1|g(x,\theta)) = \epsilon + (1-2\epsilon)\sigma(g(x,\theta))$, where $\sigma$ is the sigmoid function, $g(x,\theta)$ is the input to the final layer and $\epsilon$ is the hyperparameter.

Our method performs equally to or better than ordinary VI and other existing methods for all the datasets.

# 6 Conclusions

In this work, we proposed outlier-robust variational inference based on robust divergences. We can make our estimation robust against outliers without changing models. We also theoretically compared our proposed method and the ordinary variational inference by using the influence function. By using the influence function, we can evaluate how much outliers affect our predictions. The analysis showed that the influence of outliers is bounded in our model, but is unbounded by ordinary variational inference in many cases. Further, experiments demonstrated that our method is robust for both input and output related outliers in both regression and classification settings. In addition, our method outperforms ordinary VI on benchmark datasets.

In our future work, we would like to extend the method to be applicable to more complex models, such as time series or structured data. Another way is to combine other approximation methods such as MCMC or expectation propagation.

**Acknowledgements**

This work was supported by KAKENHI 17H00757 and JST CREST JPMJCR1403.

# References

Ayanendranath Basu, Ian R. Harris, Nils L. Hjort, and M. C. Jones. Robust and efficient estimation by minimising a density power divergence. *Biometrika*, 85 (3):549–559, 1998.

Thomas Bonald and Richard Combes. A minimax optimal algorithm for crowdsourcing. In *Advances in Neural Information Processing Systems*, pages 4355–4363, 2017.

Andrzej Cichocki, Sergio Cruces, and Shun-ichi Amari. Generalized alpha-beta divergences and their application to robust nonnegative matrix factorization. *Entropy*, 13(1):134–170, 2011.

Hironori Fujisawa and Shinto Eguchi. Robust parameter estimation with a small bias against heavy contamination. *Journal of Multivariate Analysis*, 99 (9):2053–2081, 2008.

Abhik Ghosh and Ayanendranath Basu. Robust bayes estimation using the density power divergence. *Annals of the Institute of Statistical Mathematics*, 68(2):413–437, Apr 2016.

Ryan Giordano, Tamara Broderick, and Michael Jordan. Robust inference with variational bayes. *arXiv preprint arXiv:1512.02578*, 2015.

Jose Hernandez-Lobato, Yingzhen Li, Mark Rowland, Thang Bui, Daniel Hernandez-Lobato, and Richard Turner. Black-box alpha divergence minimization. In *Proceedings of The 33rd International Conference on Machine*

*Learning*, volume 48 of *Proceedings of Machine Learning Research*, pages 1511–1520, New York, USA, 20–22 Jun 2016.

P.J. Huber and E.M. Ronchetti. *Robust Statistics*. Wiley Series in Probability and Statistics. Wiley, 2011.

Diederik P Kingma and Jimmy Ba. Adam: A method for stochastic optimization. *arXiv preprint arXiv:1412.6980*, 2014.

Pang Wei Koh and Percy Liang. Understanding black-box predictions via influence functions. In *Proceedings of the 34th International Conference on Machine Learning*, volume 70 of *Proceedings of Machine Learning Research*, pages 1885–1894, Sydney, Australia, 06–11 Aug 2017.

Oluwasanmi Koyejo and Joydeep Ghosh. *Constrained Bayesian inference for low rank multitask learning*, pages 341–350. 2013.

Yingzhen Li and Yarin Gal. Dropout inference in Bayesian neural networks with alpha-divergences. In *Proceedings of the 34th International Conference on Machine Learning*, volume 70 of *Proceedings of Machine Learning Research*, pages 2052–206, Sydney, Australia, 06–11 Aug 2017.

Yingzhen Li and Richard E Turner. Rényi divergence variational inference. In *Advances in Neural Information Processing Systems*, pages 1073–1081, 2016.

M. Lichman. UCI machine learning repository, 2013. URL http://archive.ics.uci.edu/ml.

Qiang Liu, Jian Peng, and Alexander T Ihler. Variational inference for crowdsourcing. In F. Pereira, C. J. C. Burges, L. Bottou, and K. Q. Weinberger, editors, *Advances in Neural Information Processing Systems 25*, pages 692–700. Curran Associates, Inc., 2012.

Kevin P Murphy. Machine learning: A probabilistic perspective. 2012.

Karthik Narayan, Ali Punjani, and Pieter Abbeel. Alpha-beta divergences discover micro and macro structures in data. In *Proceedings of the 32Nd International Conference on International Conference on Machine Learning*, pages 796–804, 2015.

Rajesh Ranganath, Sean Gerrish, and David Blei. Black box variational inference. In *Artificial Intelligence and Statistics*, pages 814–822, 2014.

Vikas C Raykar, Shipeng Yu, Linda H Zhao, Gerardo Hermosillo Valadez, Charles Florin, Luca Bogoni, and Linda Moy. Learning from crowds. *Journal of Machine Learning Research*, 11(Apr):1297–1322, 2010.

Wojciech Samek, Duncan Blythe, Klaus-Robert Müller, and Motoaki Kawanabe. Robust spatial filtering with beta divergence. In *Advances in Neural Information Processing Systems*, pages 1007–1015, 2013.

Aad W Van der Vaart. *Asymptotic statistics*, volume 3. Cambridge university press, 1998.

Yixin Wang, Alp Kucukelbir, and David M. Blei. Robust probabilistic modeling with Bayesian data reweighting. In *Proceedings of the 34th International Conference on Machine Learning*, volume 70 of *Proceedings of Machine Learning Research*, pages 3646–3655, Sydney, Australia, 06–11 Aug 2017.

Arnold Zellner. Optimal information processing and bayes's theorem. *The American Statistician*, 42(4):278–280, 1988.

Yuchen Zhang, Xi Chen, Dengyong Zhou, and Michael I Jordan. Spectral methods meet em: A provably optimal algorithm for crowdsourcing. *Journal of Machine Learning Research*, 17(1):3537–3580, 2016.

Jun Zhu, Ning Chen, and Eric P. Xing. Bayesian inference with posterior regularization and applications to infinite latent SVMs. *Journal of Machine Learning Research*, 15:1799–1847, 2014.

# A   $\gamma$ divergence minimization

## A.1   Unsupervised setting

In this section, we explain the $\gamma$ divergence minimization for unsupervised setting. We denote true distribution as $p^*(x)$. We denote the model by $p(x;\theta)$. We minimize the following $\gamma$ cross entropy,

$$d_\gamma(p^*(x), p(x;\theta)) = -\frac{1}{\gamma} \ln \int p^*(x) p(x;\theta)^\gamma dx + \frac{1}{1+\gamma} \ln \int p(x;\theta)^{1+\gamma} dx. \quad (27)$$

This is empirically approximated as

$$L_n(\theta) = d_\gamma(\hat{p}(x), p(x;\theta)) = -\frac{1}{\gamma} \ln \frac{1}{n} \sum_{i=1}^n p(x_i;\theta)^\gamma dx + \frac{1}{1+\gamma} \ln \int p(x;\theta)^{1+\gamma} dx. \quad (28)$$

By minimizing $L_n(\theta)$, we can obtain following estimation equation,

$$0 = -\frac{\sum_{i=1}^n p(x_i;\theta)^\gamma \frac{\partial}{\partial \theta} \ln p(x_i;\theta)}{\sum_{i=1}^n p(x_i;\theta)^\gamma} + \int \frac{p(x;\theta)^{1+\gamma}}{\int p(x;\theta)^{1+\gamma} dx} \frac{\partial}{\partial \theta} \ln p(x;\theta) dx. \quad (29)$$

This is actually weighted likelihood equation, where the weights are $\frac{p(x_i;\theta)^\gamma}{\sum_{i=1}^n p(x_i;\theta)^\gamma}$. The second term is for the unbiasedness of the estimating equation.

Actually above estimator is equivalent to minimizing following expression,

$$L'_n(\theta) = -\frac{1}{n} \sum_{i=1}^n \frac{\gamma+1}{\gamma} \frac{p(x_i|\theta)^\gamma}{\left\{ \int p(x|\theta)^{1+\gamma} dy \right\}^{\frac{\gamma}{1+\gamma}}} \quad (30)$$

In the main paper, we use $L'_n(\theta)$ as $\gamma$ cross entropy instead of using original expression. The reason is given in Appendix H.

## A.2 Supervised setting

In this section, we explain the $\gamma$ divergence minimization for the supervised setting. We denote the true distribution as $p^*(y,x) = p^*(y|x)p^*(x)$. We denote the regression model by $p(y|x;\theta)$.

Following Fujisawa and Eguchi [2008], we define the divergence between true distribution and the model by

$$D_\gamma(p^*(y|x), p(y|x;\theta)|p^*(x))$$
$$= \frac{1}{\gamma} \ln \int \left\{ \int g^*(y|x)^{1+\gamma} dy \right\}^{\frac{1}{1+\gamma}} p^*(x) dx - \frac{1}{\gamma} \ln \int \left\{ \int g^*(y|x) p(y|x;\theta)^\gamma dy / \left( \int g^*(y|x)^{1+\gamma} dy \right)^{\frac{\gamma}{1+\gamma}} \right\} p^*(x) dx. \quad (31)$$

As discussed in Fujisawa and Eguchi [2008], in the limit where $\gamma \to 0$, this divergence becomes ordinary KL divergence,

$$\lim_{\gamma \to 0} D_\gamma(p^*(y|x), p(y|x;\theta)|p^*(x)) = \int D_{\mathrm{KL}}(p^*(y|x), p(y|x;\theta)) p^*(x) dx \quad (32)$$

What we minimize is following $\gamma$ cross entropy over the distribution $p^*(x)$, Actually, minimizing $\gamma$ divergence is equivalent to minimizing the second term of Eq.(31). By empirical approximation, what we minimize is following expression,

$$L_n(\theta) = -\frac{1}{n} \sum_{i=1}^n L_n(\theta) = -\frac{1}{n} \sum_{i=1}^n \frac{p(y_i|x_i;\theta)^\gamma}{\{\int p(y|x_i;\theta)^{1+\gamma} dy\}^{\frac{\gamma}{1+\gamma}}}. \quad (33)$$

As $\gamma \to 0$, above expression goes to

$$L_n(\theta) = -\frac{1}{n} \sum_{i=1}^n \ln p(y_i|x_i;\theta). \quad (34)$$

This is ordinary KL cross entropy.

## B  $\beta$ divergence minimization

Until now, we focused on $\gamma$ divergence minimization. We can also consider supervised setup for $\beta$ divergence minimization. The empirical approximation of $\beta$ cross entropy for supervised settings is

$$L_n(\theta) = d_\beta(\hat{p}(y|x), p(y|x;\theta)|\hat{p}(x)) = -\frac{\beta+1}{\beta} \left\{ \frac{1}{n} \sum_{i=1}^n p(y_i|x_i;\theta)^\beta \right\} + \left\{ \frac{1}{n} \sum_{i=1}^n \int p(y|x_i;\theta)^{1+\beta} dy \right\}. \quad (35)$$

For the unsupervised setting, the empirical approximation of $\beta$ cross entropy is

$$L_n(\theta) = d_\beta(\hat{p}(x), p(x;\theta)) = -\frac{\beta+1}{\beta} \frac{1}{n} \sum_{i=1}^n p(x_i;\theta)^\beta + \int p(x;\theta)^{1+\beta} dx. \quad (36)$$

## C  Proof of Eq.(14) in the main paper

From the definition of KL divergence Eq.(2) in the main paper, the cross entropy can be expressed as

$$d_{\mathrm{KL}}\left(\hat{p}(x) \| p(x|\theta)\right) = D_{\mathrm{KL}}\left(\hat{p}(x) \| p(x|\theta)\right) + \mathrm{Const.} \tag{37}$$

By substituting the above expression into the definition of $L(q(\theta))$, we obtain

$$L(q(\theta)) = D_{\mathrm{KL}}(q(\theta) \| p(\theta)) + N \mathbb{E}_{q(\theta)}[D_{\mathrm{KL}}\left(\hat{p}(x) \| p(x|\theta)\right)] + \mathrm{Const.}$$

What we have to consider is

$$\arg\min_{q(\theta) \in \mathcal{P}} L(q(\theta)), \tag{38}$$

We can disregard the constant term in $L(q(\theta))$, and above optimization problem is equivalent to

$$\arg\min_{q(\theta) \in \mathcal{P}} \frac{1}{N} L(q(\theta)). \tag{39}$$

Therefore Eq.(14) is equivalent to Eq.(13)

## D  Proof of Theorem 1

The objective function is given as

$$L_\beta = \mathbb{E}_{q(\theta)}[D_\beta\left(\hat{p}(x) \| p(x|\theta)\right)] + \lambda' D_{KL}\left(q(\theta) \| p(\theta)\right) \tag{40}$$

where $\lambda'$ is the regularization constant. We optimize this with the constraint that $\int q(\theta) d\theta = 1$. We calculate using the method of variations and Lagrange multipliers, we can get the optimal $q(\theta)$ in the following way,

$$\frac{d(L_\beta + \lambda(\int q(\theta) d\theta - 1))}{dq(\theta)} = D_\beta\left(\hat{p}(x) | p(x|\theta))\right] + \lambda' \ln \frac{q(\theta)}{p(\theta)} - (1 + \lambda) = 0 \tag{41}$$

By rearranging the above expression, we can get the following relation,

$$q(\theta) \propto p(\theta) e^{-\frac{1}{\lambda'} d_\beta(\hat{p}(x) | p(x|\theta))} \tag{42}$$

If we set $\frac{1}{\lambda'} = N$ and normalize the above expression, we get the Theorem 1 in the main text,

$$q(\theta) = \frac{e^{-N d_\beta(\hat{p}(x) | p(x|\theta))} p(\theta)}{\int e^{-N d_\beta(\hat{p}(x) | p(x|\theta))} p(\theta) d\theta}. \tag{43}$$

We can get the similar expression for $\gamma$ cross entropy.

Interestingly, if we use KL cross entropy instead of $\beta$ cross entropy in the above discussion, following relation holds,

$$\begin{aligned}
q(\theta) &\propto p(\theta)e^{-\frac{1}{\lambda'}d_{KL}(\hat{p}(x)|p(x|\theta))} = p(\theta)e^{-N(-\frac{1}{N}\sum_i \ln p(x_i|\theta))} \\
&= p(\theta)\prod_i p(x_i|\theta) \\
&= p(\theta)p(D|\theta)
\end{aligned} \qquad (44)$$

The normalizing constant is

$$\int p(\theta)\prod_i p(x_i|\theta)d\theta = p(D). \qquad (45)$$

Finally, we get the optimal $q(\theta)$

$$q(\theta) = \frac{p(D|\theta)p(\theta)}{p(D)}. \qquad (46)$$

This is the posterior distribution which can be derived by Bayes' theorem.

In the above proof, we set regularization constant as $\frac{1}{\lambda'} = N$ to derive the expression. In this paper we only consider the situation that regularization constant is $\frac{1}{\lambda'} = N$ based on the similarity of Bayes' theorem. However how to choose the regularization constant should be studied further in the future because which reflects the trade off between prior information and information from data.

## E Pseudo posterior

The expression Eq.(43) is called pseudo posterior in statistics. In general, pseudo posterior is given as

$$q(\theta) = \frac{e^{-\lambda R(\theta)}p(\theta)}{\int e^{-\lambda R(\theta)}p(\theta)d\theta}. \qquad (47)$$

where $p(\theta)$ is prior and $R(\theta)$ expresses empirical risk not restricted to likelihood and not necessarily additive. The is also called Gibbs posterior and extensively studied in the field of PAC Bayes. Our $\beta$ cross entropy based pseudo posterior is

$$\begin{aligned}
q(\theta) &\propto e^{-N\{\frac{\beta+1}{\beta}\frac{1}{N}\sum_{i=1}^{N}p(x_i;\theta)^\beta + \int p(x;\theta)^{1+\beta}dx\}}p(\theta) \\
&= \left[\prod_i^N e^{l_\theta(x_i)}p(\theta)\right]
\end{aligned} \qquad (48)$$

where $l_\theta(x_i) = \frac{\beta+1}{\beta}p(x_i;\theta)^\beta - \frac{1}{N}\int p(x;\theta)^{1+\beta}dx$.

As discussed in Ghosh and Basu (2016), we can understand the intuitive meaning of above expression by comparing this expression with Eq.(44). In ordinary Bayes posterior, the prior belief is updated by likelihood $p(x_i|\theta)$ which represents the information from data $x_i$ as shown in Eq.(44). On the other hand, when using $\beta$ cross entropy, the prior belief is updated by $e^{l_\theta(x_i)}$ which has information about data $x_i$. Therefore the spirit of Bayes, that is, we update information about parameter based on training data, are inherited to this pseudo posterior.

Table 7: Cross-entropies for robust variational inference.

| | Unsupervised | Supervised |
|---|---|---|
| $d_\beta$ | $-\frac{\beta+1}{\beta}\frac{1}{N}\sum_{i=1}^{N}p(x_i|\theta)^\beta + \int p(x|\theta)^{1+\beta}dx$ | $-\frac{\beta+1}{\beta}\left\{\frac{1}{N}\sum_{i=1}^{N}p(y_i|x_i,\theta)^\beta\right\} + \left\{\frac{1}{N}\sum_{i=1}^{N}\int p(y|x_i,\theta)^{1+\beta}dy\right\}$ |
| $d_\gamma$ | $-\frac{1}{N}\frac{\gamma+1}{\gamma}\sum_{i=1}^{N}\frac{p(x_i|\theta)^\gamma}{\{\int p(x|\theta)^{1+\gamma}dx\}^{\frac{\gamma}{1+\gamma}}}$ | $-\frac{1}{N}\frac{\gamma+1}{\gamma}\sum_{i=1}^{N}\frac{p(y_i|x_i,\theta)^\gamma}{\{\int p(y|x_i,\theta)^{1+\gamma}dy\}^{\frac{\gamma}{1+\gamma}}}$ |

# F Proof of Theorem 2

We consider the situation where the distribution is expressed as

$$G_\varepsilon(x) = (1-\varepsilon)G_n(x) + \varepsilon\Delta_z(x) \qquad (49)$$

Before going to the detail, we summarize the objective function of VI and proposed method.

First, the objective function of ordinary VI is given by

$$L = D_{\mathrm{KL}}(q(\theta)\|p(\theta)) + N\mathbb{E}_{q(\theta)}\left[Nd_{\mathrm{KL}}(\hat{p}(x)\|p(x|\theta))\right]. \qquad (50)$$

In the same way, objective functions of $\beta$-VI and $\gamma$-VI are given by

$$L_\beta = D_{\mathrm{KL}}(q(\theta)\|p(\theta)) + N\mathbb{E}_{q(\theta)}\left[Nd_\beta(\hat{p}(x)\|p(x|\theta))\right], \qquad (51)$$

$$L_\gamma = D_{\mathrm{KL}}(q(\theta)\|p(\theta)) + N\mathbb{E}_{q(\theta)}\left[Nd_\gamma(\hat{p}(x)\|p(x|\theta))\right], \qquad (52)$$

where $d_\beta$ and $d_\gamma$ are summarized in Table 7. By using these expressions, we will derive the influence functions.

## F.1 Derivation of IF for ordinary VI

We start from the first order condition,

$$\begin{aligned}0 &= \left.\frac{\partial}{\partial m}L\right|_{m=m^*} \\ &= \nabla_m \mathbb{E}_{q(\theta;m^*(\epsilon))}\left[N\int dG_\epsilon(x)\ln p(x|\theta) + \ln p(\theta) - \ln q(\theta;m^*(\epsilon))\right]\end{aligned} \qquad (53)$$

We differentiate above expression with $\epsilon$, then we obtain following expression,

$$0 = \nabla_m \int d\theta \frac{\partial m^*(\epsilon)}{\partial \epsilon} \frac{\partial q}{\partial m^*(\epsilon)} \left\{ (1-\epsilon) N \int dG_n(x) \ln p(x|\theta) + \epsilon N \ln p(z|\theta) + \ln p(\theta) \right\}$$
$$+ \nabla_m \mathbb{E}_{q(\theta;m^*(\epsilon))} \left[ -N \int dG_n(x) \ln p(x|\theta) + N \ln p(z|\theta) \right]$$
$$- \nabla_m \int d\theta \frac{\partial m^*(\epsilon)}{\partial \epsilon} \frac{\partial q}{\partial m^*(\epsilon)} \ln q(\theta; m^*(\epsilon)) - \nabla_m \mathbb{E}_{q(\theta;m^*(\epsilon))} \left[ \frac{\partial m^*(\epsilon)}{\partial \epsilon} \cdot \frac{\partial \ln q}{\partial m^*(\epsilon)} \right]$$
(54)

From above expression, if we take $\epsilon \to 0$, we soon obtain following expression,

$$\frac{\partial m^*(\varepsilon)}{\partial \varepsilon} = -\left(\frac{\partial^2 L}{\partial m^2}\right)^{-1} \frac{\partial}{\partial m} \mathbb{E}_{q(\theta)} \left[ N \int dG_n(x) \ln p(x|\theta) - N \ln p(z|\theta) \right]. \quad (55)$$

Actually, this can be transformed to following expression by using the first order condition,

$$\frac{\partial m^*(\varepsilon)}{\partial \varepsilon} = \left(\frac{\partial^2 L}{\partial m^2}\right)^{-1} \frac{\partial}{\partial m} \mathbb{E}_{q(\theta)} \left[ D_{KL}(q(\theta;m)|p(\theta)) + N \ln p(z|\theta) \right]. \quad (56)$$

### F.2 Derivation of IF for $\beta$ VI

Next we consider IF for $\beta$ VI. To proceed calculation, we have to be careful that empirical approximation of $\beta$ cross entropy takes different form between unsupervised and supervised setting as shown in Eq.(36) and Eq.(35).

For the unsupervised situation, we can write the first order condition as,

$$0 = \left.\frac{\partial}{\partial m} L_\beta\right|_{m=m^*}$$
$$= \nabla_m \mathbb{E}_{q(\theta;m^*(\epsilon))} \left[ N \int dG_\epsilon(x) \frac{\beta+1}{\beta} p(x|\theta)^\beta - N \int p(x|\theta)^{1+\beta} dx + \ln p(\theta) - \ln q(\theta;m^*(\epsilon)) \right].$$
(57)

We can proceed calculation in the same way as ordinary VI. We get the following expression

$$\frac{\partial m^*(\varepsilon)}{\partial \varepsilon} = -\frac{\beta+1}{\beta} \left(\frac{\partial^2 L_\beta}{\partial m^2}\right)^{-1} \frac{\partial}{\partial m} \mathbb{E}_{q(\theta)} \left[ N \int dG_n(x) p(x|\theta)^\beta - N p(z|\theta)^\beta \right]. \quad (58)$$

Next, we consider the supervised situation. We consider the situation where the contamination is expressed as

$$G_\varepsilon(x,y) = (1-\varepsilon) G_n(x,y) + \varepsilon \Delta_{z=(x',y')}(x,y) \quad (59)$$

The first order condition is,

$$0 = \frac{\partial}{\partial m} L_\beta \bigg|_{m=m^*}$$
$$= \nabla_m \mathbb{E}_{q(\theta;m^*(\epsilon))} \left[ N \int dG_\epsilon(x,y) \frac{\beta+1}{\beta} p(y|x,\theta)^\beta - N \int dG_\epsilon(x) \left\{ \int p(y|x,\theta)^{1+\beta} dy \right\} + \ln p(\theta) - \ln q(\theta; m^*(\epsilon)) \right] \quad (60)$$

We can proceed the calculation and derive the influence function as follows,

$$\frac{\partial m^*(\varepsilon)}{\partial \varepsilon} = -N \left( \frac{\partial^2 L_\beta}{\partial m^2} \right)^{-1} \frac{\partial}{\partial m} \mathbb{E}_{q(\theta)} \left[ \frac{\beta+1}{\beta} \left( \int dG_n(y,x) p(y|x,\theta)^\beta - p(y'|x',\theta)^\beta \right) \right]$$
$$+ N \left( \frac{\partial^2 L_\beta}{\partial m^2} \right)^{-1} \frac{\partial}{\partial m} \mathbb{E}_{q(\theta)} \left[ \int dG_n(x) \left( \int p(y|x,\theta)^{1+\beta} dy \right) - \int p(y|x',\theta)^{1+\beta} dy \right]. \quad (61)$$

If we take the limit $\beta$ to 0, the above expression reduced to IF of ordinary VI.

### F.3 Derivation of IF for $\gamma$ VI

We can derive IF for $\gamma$ VI in the same way as $\beta$ VI.

For simplicity, we focus on the transformed cross entropy, which is given Eq.(34). For unsupervised situation, the first order condition is given by,

$$0 = \frac{\partial}{\partial m} L_\gamma \bigg|_{m=m^*}$$
$$= \nabla_m \mathbb{E}_{q(\theta;m^*(\epsilon))} \left[ N \int dG_\epsilon(x) \frac{p(x|\theta)^\gamma}{\left\{ \int p(x|\theta)^{1+\gamma} dx \right\}^{\frac{\gamma}{1+\gamma}}} + \ln p(\theta) - \ln q(\theta; m^*(\epsilon)) \right]. \quad (62)$$

In the same way as $\beta$ VI, we can get the IF of $\gamma$ VI for unsupervised setting as,

$$\frac{\partial m^*(\varepsilon)}{\partial \varepsilon} = -\left( \frac{\partial^2 L_\gamma}{\partial m^2} \right)^{-1} \frac{\partial}{\partial m} \mathbb{E}_{q(\theta)} \left[ N \frac{\int dG_n(x) p(x|\theta)^\gamma - p(z|\theta)^\gamma}{\left\{ \int p(x|\theta)^{1+\gamma} dx \right\}^{\frac{\gamma}{1+\gamma}}} \right]. \quad (63)$$

For supervised situation, the first order condition is give by,

$$0 = \frac{\partial}{\partial m} L_\gamma \bigg|_{m=m^*}$$
$$= \nabla_m \mathbb{E}_{q(\theta;m^*(\epsilon))} \left[ N \int dG_\epsilon(x,y) \frac{p(y|x,\theta)^\gamma}{\left\{ \int p(y|x,\theta)^{1+\gamma} dy \right\}^{\frac{\gamma}{1+\gamma}}} + \ln p(\theta) - \ln q(\theta; m^*(\epsilon)) \right]. \quad (64)$$

In the same way as $\beta$ VI, we can get the IF of $\gamma$ VI for supervised setting as,

$$\frac{\partial m^*(\varepsilon)}{\partial \varepsilon} = -N \left( \frac{\partial^2 L_\gamma}{\partial m^2} \right)^{-1} \frac{\partial}{\partial m} \mathbb{E}_{q(\theta)} \left[ \int dG_n(x,y) \frac{p(y|x,\theta)^\gamma}{\left\{ \int p(y|x,\theta)^{1+\gamma} dy \right\}^{\frac{\gamma}{1+\gamma}}} - \frac{p(y'|x',\theta)^\gamma}{\left\{ \int p(y|x',\theta)^{1+\gamma} dy \right\}^{\frac{\gamma}{1+\gamma}}} \right]. \quad (65)$$

## G  Other aspects of analysis based on influence function

In the above sections, we considered that outliers are added to the original training dataset. We can consider a other type of contamination, such as training data itself is perturbed, that is, a training point $z = (x, y)$ is perturbed to $z_\epsilon = (x + \epsilon, y)$(Koh and Liang (2017)). We call this type of data contamination as data perturbation. As for data perturbation, following relation holds,

When we consider data perturbation for a training data, IF of ordinary VI is given by

$$\frac{\partial m^*(\varepsilon)}{\partial \varepsilon} = -\left(\frac{\partial^2 L}{\partial m^2}\right)^{-1} \frac{\partial}{\partial m} \mathbb{E}_{q(\theta)}\left[\frac{\partial}{\partial x} \ln p(z|\theta)\right]. \quad (66)$$

IF of $\beta$ divergence based VI is given by

$$\frac{\partial m^*(\varepsilon)}{\partial \varepsilon} = -\left(\frac{\partial^2 L_\beta}{\partial m^2}\right)^{-1} \frac{\partial}{\partial m} \mathbb{E}_{q(\theta)}\left[\frac{\partial}{\partial x} p(z|\theta)^\beta\right]. \quad (67)$$

## H  Another type of $\gamma$ VI

In the main paper, we used the transformed $\gamma$ cross entropy, which is given in Eq.(33). The reason we used the transformed cross entropy instead of original expression is that we can interpret the pseudo posterior when using the transformed cross entropy much easily than when using original cross entropy.

In the same way Eq.(68), we can derive the pseudo posterior using transformed cross entropy,

$$q(\theta) \propto e^{N\frac{\gamma+1}{\gamma}\frac{1}{N}\sum_{i=1}^N \frac{p(x_i|\theta)^\gamma}{\{\int p(x|\theta)^{1+\gamma}dy\}^{\frac{\gamma}{1+\gamma}}}} p(\theta)$$
$$= \left[\prod_i^N e^{l_\theta(x_i)} p(\theta)\right] \quad (68)$$

where $l_\theta(x_i) = \frac{\gamma+1}{\gamma}\frac{p(x_i|\theta)^\gamma}{\{\int p(x|\theta)^{1+\gamma}dy\}^{\frac{\gamma}{1+\gamma}}}$. In this formulation, it is easy to consider that the information of data $x_i$ is utilized to update the prior information through $e^{l_\theta(x_i)}$.

However, when using original cross entropy, such interpretation cannot be done because the pseudo posterior is given by,

$$q(\theta) \propto e^{N(\frac{1}{\gamma}\ln\frac{1}{N}\sum_i^N p(x_i|\theta)^\gamma dx - \frac{1}{1+\gamma}\ln\int p(x|\theta)^{1+\gamma}dx)} p(\theta) \quad (69)$$

and since the summation is not located in the front, this pseudo posterior has not additivity. Therefore it is difficult to understand how each training data $x_i$ contributes to update the parameter. Moreover it is not straight forward to apply stochastic variational inference framework. Accordingly, we decided to use the transformed cross entropy.

Even thought the interpretation is difficult we can dirive IF in the same way as we discussed. For unsupervised situation, the first order condition is given by

$$0 = \left.\frac{\partial}{\partial m} L_\gamma \right|_{m=m^*}$$
$$= \nabla_m \mathbb{E}_{q(\theta; m^*(\epsilon))} \left[ \frac{N}{\gamma} \ln \int dG_\epsilon(x) p(x|\theta)^\gamma dx - \frac{N}{1+\gamma} \ln \int p(x|\theta)^{1+\gamma} dx + \ln p(\theta) - \ln q(\theta; m^*(\epsilon)) \right]. \tag{70}$$

In the same way as $\beta$ VI, we can get the IF of $\gamma$ VI of original cross entropy for unsupervised setting as,

$$\frac{\partial m^*(\varepsilon)}{\partial \varepsilon} = -\frac{N}{\gamma} \left(\frac{\partial^2 L_\gamma}{\partial m^2}\right)^{-1} \frac{\partial}{\partial m} \mathbb{E}_{q(\theta)} \left[ \frac{\int dG_n(x) p(x|\theta)^\gamma - N p(z|\theta)^\gamma}{\int dG_n(x) p(x|\theta)^\gamma} \right]. \tag{71}$$

For supervised situation, we can derive in the same way.

# I  Discussion of Influence function

In this section, we describe the detail discussion of influence function's behavior when using a neural net model for the regression and the classification with logistic loss.

We focus on the influence function of the variational parameter in the approximate posterior distribution. We use mean-field variational inference and Gaussian distribution for approximate posterior. $q(\theta)$ denote the approximate posterior. Since Gaussian distribution is a member of an exponential family, we can parametrize it by its mean value $m$. In the case of Gaussian distribution, $m = \{\mathbb{E}[\theta], \mathbb{E}[\theta^2]\}$. We can parametrize variational posterior as $q(\theta|m)$. Thus we only analyze the influence function of $m = \mathbb{E}[\theta]$ in this section and $m$ indicates the $m = \mathbb{E}[\theta]$ not $\mathbb{E}[\theta^2]$.

Let us start ordinary variational inference. In Eq.(66), we especially focus on the term, $\frac{\partial}{\partial m} \mathbb{E}_{q(\theta|m)} [\ln p(y|\theta)]$, because this is the only term that is related to outlier. If we assume that approximate posterior is an Gaussian distribution, we can transform this term in the following way,

$$\frac{\partial}{\partial m} \mathbb{E}_{q(\theta|m)} [\ln p(y|\theta)] = \frac{\partial}{\partial m} \left\{ \int q(\theta|m) \ln p(y|\theta) d\theta \right\}$$
$$= \int \frac{\partial q(\theta|m)}{\partial m} \ln p(y|\theta) d\theta$$
$$= -\int q(\theta|m) \frac{\partial}{\partial \theta} \ln p(y|\theta) d\theta$$
$$= -E_{q(\theta|m)} \left[ \frac{\partial}{\partial \theta} \ln p(y|\theta) \right] \tag{72}$$

, where We used partial integration for the second line to third line. and also used the following relation which holds for Gaussian distribution

$$\frac{\partial q\left(\theta|m\right)}{\partial m} = \frac{\partial q\left(\theta|m\right)}{\partial \theta}. \tag{73}$$

This relation also holds for the Student-T

From above expression, it is clear that studying the behavior of $\frac{\partial}{\partial \theta} \ln p\left(y|\theta\right)$ is crucial for analyzing IF. In this case, the behavior of IF in this expression is similar to that of maximum likelihood. The related discussion are shown in AppendixJ

## I.1 Regression

In this subsection, we consider the regression problem by a neural network. We denote the input to the final layer as $f_\theta(x) \sim p(f|x,\theta)$, where $x$ is the input and $\theta$s are random variables which obeys approximate posterior $q(\theta|m)$.

We consider the output layer as Gaussian distribution as $p(y|f_\theta(x)) = N(y|f_\theta(x), I)$. From above discussion, what we have to consider is $\frac{\partial}{\partial \theta} \ln p\left(y|f_\theta(x)\right)$.

We denote input related outlier as $x_o$, that means $x_o$ does not follow the same distribution as other regular training dataset. Also, we denote the output related outlier as $y_o$ that it does not follow the same observation noise as other training dataset.

**Output related outlier**

Since we consider the model that output layer is Gaussian distribution, following relation holds for IF of ordinary VI,

$$\frac{\partial}{\partial \theta} \ln p\left(y_o|f_\theta(x_o)\right) \propto (y_o - f_\theta(x_o)) \frac{\partial f_\theta(x_o)}{\partial \theta}. \tag{74}$$

We can see that this term does not bounded when $y_o \to \pm\infty$. And thus IF of ordinary VI is unbounded as output related outlier become large.

As for the $\beta$ divergence, we have to treat Eq.(61). Fortunately, when we use Gaussian distribution for output layer, the second term in the bracket of Eq.(61) will be constant by the analytical integration, and thus its derivative will be zero. Therefore the output related term is only the first term. Thanks to this property, the denominator of Eq.(65) will also be a constant. Therefore IF of $\beta$ VI and $\gamma$ VI behaves in the same way. Therefore, we only consider $\beta$ VI for the regression. We get the following expression,

$$\frac{\partial}{\partial \theta} p\left(y_o|f_\theta(x_o)\right)^\beta \propto e^{-\frac{\beta}{2}(y_o-f_\theta(x_o))^2} (y_o - f_\theta(x_o)) \frac{\partial f_\theta(x_o)}{\partial \theta}$$
$$= \frac{(y_o - f_\theta(x_o))}{e^{\frac{\beta}{2}(y_o-f_\theta(x_o))^2}} \frac{\partial f_\theta(x_o)}{\partial \theta} \tag{75}$$

From this expression, we can see that IF of $\beta$ VI is bounded because Eq.(75) goes to 0 as $y_o \to \pm\infty$. This means that the influence of this contamination will become zero. This is the desired property for robust estimation.

**Input related outlier**

Next, we consider input related outlier. We consider whether Eq.(74) and Eq.(75) are bounded or not when $x_o \to \pm\infty$.

To proceed the analysis, we have to specify models. We start from the most simple case, $f_\theta(x_o) = W_1 x_o + b_1$, where $\theta = \{W_1, b_1\}$. This is the simple linear regression. In this case $\frac{\partial f_\theta(x_o)}{\partial W_1} = x_o$ and $\frac{\partial f_\theta(x_o)}{\partial b_1} = 1$. When $x_o \to \pm\infty$, $f_\theta(x_o) \to \pm\infty$.

From these fact, we can soon find that Eq.(74) is unbouded. As for Eq.(75), the exponential function in the denominator of Eq.(75) plays a crucial role. Thanks to this exponential function,

$$\frac{\partial}{\partial W_1} p(y_o|f_\theta(x_o))^\beta \propto \frac{(y_o - f_\theta(x_o))}{e^{\frac{\beta}{2}(y_o - f_\theta(x_o))^2}} x_o \xrightarrow[x_o \to \infty]{} 0 \tag{76}$$

From these facts, ordinary VI is not robust against input related outliers, however $\beta$ VI is robust.

Next we consider the situation that there is a hidden layer, that is $f_\theta(x_o) = W_2(W_1 x_o + b_1) + b_2$, where $\theta = \{W_1, b_1, W_2, b_2\}$. At this point, we do not consider activation function. Following relations hold,

$$\frac{\partial}{\partial W_1} f_\theta(x_o) = W_2 x_o, \quad \frac{\partial}{\partial W_2} f_\theta(x_o) = W_1 x_o + b_1 \tag{77}$$

From these relations, the behavior of IF in the case of $x_o \to \pm\infty$ is actually as same as the case where there is no hidden layers. Therefore, IF of input related outlier is bounded in $\beta$ VI and that is unbounded in ordinary VI. Even if we add more layers the situation does not change in this situation where no activation exists.

Next, we consider the situation that there exists activation function. We consider $relu$ and $tanh$ as activation function. In the situation that there is only one hidden layers, $f_\theta(x_o) = W_2(relu(W_1 x_o + b_1)) + b_2$,

$$\frac{\partial f_\theta(x_o)}{\partial W_2} = relu(W_1 x_o + b_1), \quad \frac{\partial f_\theta(x_o)}{\partial W_1} = \begin{cases} W_2 x_o, & W_1 x_o + b_1 \geq 0 \\ 0, & W_1 x_o + b_1 < 0, \end{cases} \tag{78}$$

Actually, this is almost the same situation as when there are no activation functions, because there remains possibility that IF will diverge in ordinary VI, while IF in $\beta$ VI is bounded.

When we use $tanh$ as a activation function, $f_\theta(x_o) = W_2 tanh(W_1 x_o + b_1) + b_2$,

$$\frac{\partial f_\theta(x_o)}{\partial W_1} = \frac{W_2 x_o}{cosh^2(W_1 x_o + b_1)} \xrightarrow[x_o \to \infty]{} 0 \quad (79)$$

The limit of above expression can be easily understand from Fig.5. From this expression, we can understand IF of $W_1$ is bounded in both ordinary estimator and $\beta$ estimator, when we consider the model, $f_\theta(x_o) = tanh(W_1 x_o + b_1)$. As for $W_2$,

$$\frac{\partial f_\theta(x_o)}{\partial W_2} = tanh(W_1 x_o + b_1) \quad (80)$$

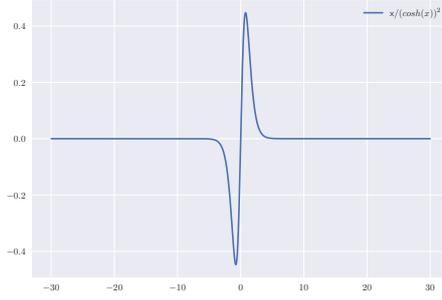

Figure 5: Behavior of $\frac{x}{\cosh^2 x}$

In this expression, even if input related outlier goes to infinity, the maximum of above expression is 1. Accordingly, the IF of $W_2$ is bounded in any case. And thus IF of both ordinary VI and $\beta$ VI is bounded when we use $tanh$ activation function.

Up to now, we have seen the model which has a hidden model. The same discussion can be held for the model which has much more hidden layers. If we add layers, above discussion holds and there remains possibility that IF using relu in ordinary VI will diverge.

We can say that ordinary VI is not robust to output related outliers and input related outliers. The exception is that using tanh activation function makes the IF of ordinary VI bounded. In $\beta$ VI, the IF of parameters are always bounded.

**Using Student-T output layer**

We additionaly consider the property of Student-t loss in terms of IF. When we denote degree of freedom as $\nu$, and the variance as $\sigma^2$, following relation holds,

$$\frac{\partial}{\partial \theta} \ln p(y_o | f_\theta(x_o)) \propto \frac{(y_o - f_\theta(x_o))}{\nu \sigma^2 + (y_o - f_\theta(x_o))^2} \frac{\partial f_\theta(x_o)}{\partial \theta} \quad (81)$$

By comparing Eq.(81) with Eq.(74) and Eq.(75), we can confirm that the behavior of IF in the case of Student-t loss in ordinary VI is similar to Gaussian loss model in $\beta$ VI. First, consider output related outlier,

$$\frac{\partial}{\partial \theta} \ln p(y_o|f_\theta(x_o)) \xrightarrow[y_o \to \infty]{} 0 \tag{82}$$

From above expression, we can find that Student-T loss is robust to output related outlier. This is the desiring property of Student-T.

Next consider input related outlier. We consider the model, $f_\theta(x_o) = W_1 x_o + b_1$, where $\theta = \{W_1, b_1\}$

$$\begin{aligned}
\frac{\partial}{\partial W_1} \ln p(y_o|f_\theta(x_o)) &\propto \frac{(y_o - f_\theta(x_o))}{\nu\sigma^2 + (y_o - f_\theta(x_o))^2} x_o \\
&= \frac{(y_o - f_\theta(x_o))^2}{\nu\sigma^2 + (y_o - f_\theta(x_o))^2} \frac{x_o}{y_o - f_\theta(x_o)} \\
&= \frac{(y_o - f_\theta(x_o))^2}{\nu\sigma^2 + (y_o - f_\theta(x_o))^2} \frac{f_\theta(x_o) - b_1}{W_1(y_o - f_\theta(x_o))} \\
&\xrightarrow[x_o \to \infty]{} -W_1^{-1}
\end{aligned} \tag{83}$$

This is an interesting result that in $\beta$ VI, the effect of input related outlier goes to 0 in the limit, on the other hand for Student-t loss, the IF is bounded but finite value remains.

Although the finite value remains in IF when using Student-T loss and its value is $W_1$, the value is considerably small. Therefore we can disregard the remained influence of Student-T loss in practice.

### I.2 Classification

In this section, we consider the classification problem. We focus on the binary classification, and output $y$ can take +1 or 0. We only consider the input related outlier for the limit discussion because the influence caused by label misspecification is always bounded.

As the model, we consider the logistic regression model,

$$p(y|f_\theta(x)) = f_\theta(x)^y (1 - f_\theta(x))^{(1-y)} \tag{84}$$

where

$$f_\theta(x) = \frac{1}{1 + e^{-g_\theta(x)}} \tag{85}$$

where $g_\theta(x)$ is input to sigmoid function. We consider a neural net for $g_\theta(x)$ later.

We first assume $g_\theta(x) = Wx + b$, then $\frac{\partial g}{\partial W} = x$ and $\frac{\partial g}{\partial b} = 1$. We assume prior and posterior distribution of W and b are Gaussian distributions. For IF analysis, we first consider the first term of Eq.(61) and only consider outlier related term inside it. To proceed the calculation, we can use the relation Eq.(72), and what we have to analyze is

$$\frac{\partial}{\partial \theta} \ln p(y|f_\theta(x)) = \frac{\partial}{\partial \theta} \left( y \ln f_\theta(x) + (1-y) \ln(1 - f_\theta(x)) \right)$$
$$= -y(1-f)\frac{\partial g}{\partial \theta} + (1-y)f\frac{\partial g}{\partial \theta} \tag{86}$$

Let us consider, for example $y = +1$

$$\frac{\partial}{\partial \theta} \ln p(y=+1|f_\theta(x)) = \frac{1}{1 + e^{g_\theta(x)}} \frac{\partial g}{\partial \theta} \tag{87}$$

As for $\theta = b$, this is always bounded. As for $\theta = W$,

$$\frac{\partial}{\partial W} \ln p(y=+1|f_\theta(x)) = \frac{1}{1 + e^{Wx+b}} x \tag{88}$$

In above expression, if we take limit $x \to +\infty$, and if $Wx \to -\infty$, above expression can diverge. If $Wx \to \infty$ when $x \to +\infty$, above expression goes to 0. From this observation, it is clear that there is a possibility that IF for input related outlier diverges in a logistic regression for ordinary VI.

As for $\beta$ VI, we have to consider the following term,

$$p(y=+1|f_\theta(x))^\beta \frac{\partial}{\partial \theta} \ln p(y=+1|f_\theta(x)) = \frac{1}{(1 + e^{-g_\theta(x)})^\beta} \frac{1}{1 + e^{g_\theta(x)}} \frac{\partial g}{\partial \theta} \tag{89}$$

This expression converges to 0 when $x_o \to \pm\infty$. In addition, we have to consider the behavior of the second term in Eq.(61) for analysis of IF, which is vanish in the regression situation. The second term of Eq.(61) can be written as

$$\left(\frac{\partial^2 L_\beta}{\partial m^2}\right)^{-1} \frac{\partial}{\partial m} \mathbb{E}_{q(\theta)} \left[ N \int p(y|x_o, \theta)^{1+\beta} dy \right]$$
$$= N \left(\frac{\partial^2 L_\beta}{\partial m^2}\right)^{-1} \frac{\partial}{\partial m} \mathbb{E}_{q(\theta)} \left[ f_\theta(x_o)^{1+\beta} + (1 - f_\theta(x_o))^{1+\beta} \right] \tag{90}$$

To proceed the analysis, we can use the relation Eq.(72). Since the inverse of hessian matrix is not related to outlier, what we have to consider is

$$\int d\theta q(\theta) \frac{\partial}{\partial \theta} f_\theta(x_o)^{1+\beta} + \frac{\partial}{\partial \theta}(1 - f_\theta(x_o))^{1+\beta}$$
$$= -\int d\theta q(\theta) \left( f_\theta(x_o)^{1+\beta}(1 - f_\theta(x_o))\frac{\partial g}{\partial \theta} + (1 - f_\theta(x_o))^{1+\beta} f_\theta(x_o)\frac{\partial g}{\partial \theta} \right)$$
$$= -\int d\theta q(\theta) \left\{ (1 - f_\theta(x_o))^\beta + f_\theta(x_o)^\beta \right\} (1 - f_\theta(x_o)) f_\theta(x_o) \frac{\partial g}{\partial \theta} \tag{91}$$

Since in the logistic regression situation, $f_\theta$ is bounded under from 0 to 1, the term $(1 - f_\theta(x_o))^\beta + f_\theta(x_o)^\beta$ cannot goes to zero. Therefore, what we have to consider is the term $(1 - f_\theta(x_o))f_\theta(x_o)\frac{\partial g}{\partial \theta}$.

$$(1 - f_\theta(x_o))f_\theta(x_o)\frac{\partial g}{\partial \theta} = \frac{1}{1 + e^{g_\theta}} \frac{1}{1 + e^{-g_\theta}} \frac{\partial g}{\partial \theta}$$
$$\xrightarrow[x_o \to \infty]{} 0 \quad (92)$$

Therefore, in the limit discussion, we do not have to consider the behavior of second term of Eq.(61). The behavior of IF is determined by the first term of Eq.(61). Accordingly, IF of logistic regression when using $\beta$ VI is bounded.

Consider the case where there exists activation functions such as *relu* or *tanh*. Since we do not use activation function for the final layer, the IF of logistic regression using *relu* activation function is not bounded when using ordinary VI because there remains a possibility that $g_\theta(x) \to -\infty$ as $x \to \pm\infty$. In such a case, our analyzing term can diverge. When using *tanh* activation function, as we discussed in regression setup, IF are always bounded.

Accordingly, our conclusion is that for the logistic regression, *relu* activation function is not robust against input related outliers when using ordinal VI, while *tanh* activation function is robust. As for $\beta$ VI, it is apparent from Eq.(89) and Eq.(92) that IF is bounded for both relu and tanh even using neural net.

Next, we consider the case of $\gamma$ VI, and what we have to analyze is the second term of Eq.(65). To proceed the analysis, we can use the relation Eq.(72). Since the inverse of hessian matrix is not related to outlier, what we have to analyze is,

$$\int d\theta q(\theta) \frac{\partial}{\partial \theta} \frac{p(y'|x')^\gamma}{\{\int p(y|x',\theta)^{1+\gamma} dy\}^{\frac{\gamma}{1+\gamma}}}$$
$$= \int d\theta q(\theta) \frac{\{\int p(y|x',\theta)^{1+\gamma} dy\}^{\frac{\gamma}{1+\gamma}} \frac{\partial}{\partial \theta} p(y'|x')^\gamma - p(y'|x')^\gamma \frac{\partial}{\partial \theta} \{\int p(y|x',\theta)^{1+\gamma} dy\}^{\frac{\gamma}{1+\gamma}}}{\{\int p(y|x',\theta)^{1+\gamma} dy\}^{\frac{2\gamma}{1+\gamma}}}.$$
(93)

In the above expression, what we have to consider is the numerator. The analysis of first term can be done in the same way as Eq.(89). Therefore it is bounded for both relu and tanh. The second term can be analyzed in the same way as Eq.(91), we do not have to consider it in the limit. From above discussion, the behavior of IF for $\gamma$ VI is the same as that for $\beta$ VI in the limit, accordingly, it is bounded even if using *relu* activation function.

## J  Analysis based on influence function under no model assumption

Let us compare the behavior of IF of ordinal VI and our proposed methods intuitively. First we consider ordinary VI. In Eq.(55), since the term which depends on contamination is $\frac{\partial}{\partial m}\mathbb{E}_{q(\theta)}\left[\ln p\left(z|\theta\right)\right]$, we focus on that term. It is difficult to deal with this expression directly, we focus on typical value of $q\left(\theta\right)$, the mean value $m$. In such a simplified situation, what we have to consider is following expression.

$$\frac{\partial}{\partial m}\ln p\left(z;m\right) \tag{94}$$

This is the ordinary maximum likelihood estimator.

Let us consider the unsupervised $\beta$ VI. What we consider is,

$$\frac{\partial}{\partial m}\left(p\left(z;m\right)\right)^{\beta} = \left(p\left(z;m\right)\right)^{\beta}\frac{\partial}{\partial m}\ln p\left(z;m\right) \tag{95}$$

To proceed the analysis, it is necessary to specify a model $p(z;\theta)$, otherwise we cannot evaluate differentiation. Here for intuitive analysis, we simply consider the behavior of $\ln p(z;m)$ and $p(z;m)^{\beta}\ln p(z;m)$, and in the case of $z$ is outlier, that is $p(z;m)$ is quite small.

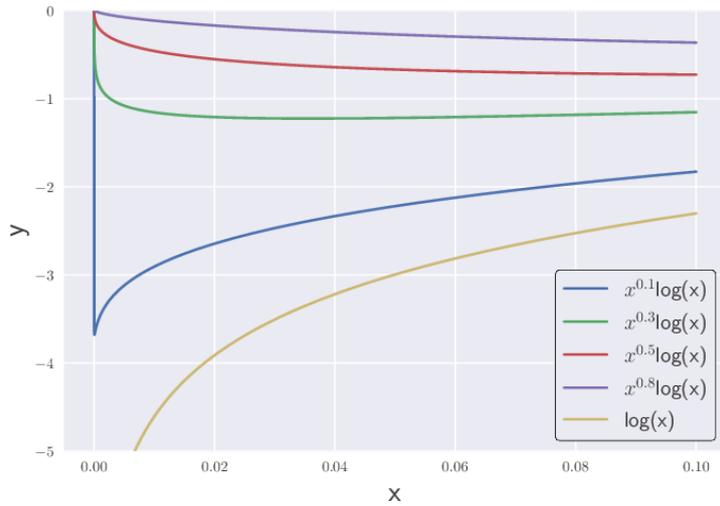

Figure 6: Behavior of $y = \log x$ and $y = x^{\beta}\log x$. As x become small, $y = \log x$ diverges to $-\infty$, on the other hand $y = x^{\beta}\log x$ is bounded.

Fig.1 shows that $\ln p(z;m)$ is unbounded, on the other hand $p(z;m)^{\beta}\ln p(z;m)$ is bounded. This means that $\beta$ divergence VI is robust to outliers.

## K  Comparison of $\beta$ VI and $\gamma$ VI

In this section, we compare the proposed $\beta$ VI and $\gamma$ VI theoretically. Although $\beta$ VI and $\gamma$ VI have robustness in based on the influence function analysis, their robustness property have significant difference if the proportion of contamination is large. If the proportion of contamination is large the assumption of discussion of influence function does not hold because we assumed that the $\epsilon$ is near zero to derive the influence function.

If the proportion of contamination is not small, other kinds of discussion is needed. Such a discussion is given in in Fujisawa and Eguchi [2008], therefore we review it and use it for our variational objectives.

Following the notation in Fujisawa and Eguchi [2008], $g(x)$ denotes the contaminated probability density function,

$$g(x) = (1 - \epsilon)f(x) + \epsilon\delta(x), \qquad (96)$$

where $f(x)$ is the underlying true probability density function, $\delta(x)$ denotes the contamination probability density function, and $\epsilon$ is the contamination proportion.

We assume that when a data point $x^*$ is an outlier $f(x^*)$ is sufficiently small. We express this assumption by saying that the following quantity is sufficiently small for an appropriate large $\gamma_0 > 0$,

$$\nu_f = \left\{\int \delta(x)f(x)^{\gamma_0}dx\right\}^{1/\gamma_0}. \qquad (97)$$

This means that $\delta(x)$ exists on the tail of $f(x)$. If $\delta(x)$ is the Dirac function at $x^*$, $\nu_f = f(x^*)$, and above assumption simply means when a data point $x^*$ is an outlier $f(x^*)$ is sufficiently small.

Under this assumption, following lemma and theorem holds (this is lemma 3.1 and theorem 3.2 in Fujisawa and Eguchi [2008]) that

**Lemma 1** *Suppose that the positive function $h$ satisfies the above assumption, where $f$ is replaced by $h$. It then holds*

$$\begin{aligned} d_\gamma(g, h) &= d_\gamma((1-\epsilon)f, h) + O(\epsilon\nu_h^\gamma) \\ &= d_\gamma(f, h) - \frac{1}{\gamma}\log(1-\epsilon) + O(\epsilon\nu^\gamma) \end{aligned} \qquad (98)$$

**Theorem 3** *Suppose that the positive function $h$ satisfies the above assumption, where $f$ is replaced by $h$. Let $\nu = \max\{\nu_f, \nu_h\}$. Then, the Pythagorean relation among $g$, $f$, and $h$ approximately holds:*

$$\Delta(g, f, h) = D_\gamma(g, h) - D_\gamma(g, f) - D_\gamma(f, h) = O(\epsilon\nu^\gamma) \qquad (99)$$

This theorem means that the minimizing divergence from the model $h$ to contaminated density $g$ is approximately equivalent to minimizing the divergence $h$ to true distribution $f$ and its order of error is given by $O(\epsilon\nu^\gamma)$.

Recall that the objective function of our proposed is given by

$$L_\gamma(q(\theta)) = \int q(\theta)\left(Nd_\gamma\left(g(x)\|p(x|\theta)\right)\right)d\theta + D_{\mathrm{KL}}(q(\theta)\|p(\theta)), \qquad (100)$$

where $g(x)$ is the contaminated distribution and $p(x|\theta)$ is the model we prepared. By using the Pythagorean relation, we can rewrite the above expression in the following way by using the true underlying distribution,

$$L_\gamma(q(\theta)) = \int q(\theta) \left( N d_\gamma(f(x)\|p(x|\theta)) - \frac{1}{\gamma}\log(1-\epsilon) + O(\epsilon\nu^\gamma) \right) d\theta + D_{\mathrm{KL}}(q(\theta)\|p(\theta)). \tag{101}$$

This equation means that by using the $\gamma$ cross entropy, we can utilize the $\gamma$ cross entropy between true distribution to our model. We optimized the objective function by using the black-box variational inference method and optimize the variational parameters by gradient decent, and thus the constant terms inside the integral are neglected.

This relation is obtained under the assumption of Eq. (97). The assumption is not the assumption that we used in the influence function that contamination proportion of $\epsilon$ is small. Therefore even if the contamination proportion is large, we can obtain the Actually, the robustness of $\beta$ divergence is assured by the influence function (Basu et al. [1998]) and thus it is not guaranteed if the contamination proportion is not sufficiently small. Following this observation, $\gamma$ divergence based method is superior to $\beta$ divergence method.

## L  Experimental detail and results

In the numerical experiment, all the prior distributions are standard Gaussian distributions. We used mean-field variational inference and we used Gaussian distributions as approximate posteriors.

### L.1  Toy experiment

For the regression task, we generated the toy data by using (x,y) by $y \sim w^\top x + \epsilon$, where $w^\top = (-0.5, -0.1)$, $x \sim N(0,1)$ and $\epsilon \sim N(0, 0.1)$. We generated 1000 data points. Outliers are generated by $x \sim N(-15, 1)$ which we considered the measurement error. We generate 24 outliers, which is 2.4% of the regular dataset. We used the linear regression, $p(y|x) = N(y|f_\theta(x), 1), f_\theta(x) = Wx + b$.

For the binary classification, the toy data are generated with the probability $p(x|y = +1) = N(x|\mu_1, \sigma_1)$, $p(x|y = -1) = N(x|\mu_2, \sigma_2)$, where $\mu_1^\top = (-1, -1)$, $\mu_2^\top = (1, 1)$, $\sigma_1 = I$, $\sigma_2 = 4I$, where $I$ is identity matrix. We generate 1000 data for each class, and in total 2000 regular points. As outliers we generate 30 outliers by using $p(x|y = +1) = N(x|\mu_o, \sigma_o)$, where $\mu_o^\top = (7, 0)$, $\sigma_2 = 0.1I$. The outliers are shown by stars in the picture in the main paper. For binary classification, we use logistic regression, where $p(y = +1|x) = logit(f_\theta(x)), f_\theta(x) = Wx + b$. We prepare priors and posteriors in the same way with binary classification.

The performance of ordinary VI estimation and our proposing methods are shown in Table.8. Apparently, the performance of ordinary VI significantly deteriorates when adding outliers. On the other hand, the performance of our proposing method is not affected by outliers.

Table 8: RMSE of VI and $\beta=0.1$ VI for toy data.

| Outliers | KL(Gaussian) | $\beta = 0.1$(Gaussian) |
|---|---|---|
| No outliers | 0.01 | 0.01 |
| Outlier exists | 0.69 | 0.01 |

Table 9: Accuracy of VI and $\beta=0.4$ VI for toy data.

| Outliers | KL(logistic) | $\beta = 0.4$(logistic) |
|---|---|---|
| No outliers | 0.97 | 0.97 |
| Outlier exists | 0.95 | 0.97 |

The illustrative results are shown in the main text. We also show the performance on this toy experiment in Table 8 and Table 9. Those tables show that ordinary VI is heavily affected by outliers, while our method is not affected so much.

## L.2 Influence function

Based on the discussion of Appendix I, the dominant term in IF of $\gamma$ VI behaves similarly with $\beta$ VI, therefore we also expected that the perturbation of predictive distribution by outliers in $\gamma$ VI behaves in the same way as $\beta$ VI. And thus, we numerically studied the perturbation of predictive distribution only about ordinary VI and $\beta$ VI. In each calculation, we used 200MC samples to get stable curves.

**Regression**

We investigated three cases where there is only input related outliers and only output related outliers and both outliers exist.

For an easy visualization and computational savings, we only contaminated the chosen single feature of the input. Since inputs have four dimensional features, $x \in \mathbb{R}^4$, we chose the first feature $x_1$ to contaminate. To investigate the how predictive distribution depends on the contamination of the input, we chose randomly a single data point from the training data and moved the value of the first feature of the chosen data from $-\infty$ to $\infty$.

For the output related outlier setting, we chose randomly a single data point from the training data and moved the output value of chosen data from $-\infty$ to $\infty$.

For both input and output related outlier setting, we chose randomly a single data point from the training data and moved the first feature of the input and the output of chosen data from $-\infty$ to $\infty$.

In the main paper, the figure of input and output related outlier settings are shown. Here we show the both the input and output related outlier situation and the graph is the case when the first feature of the input and the output value increase simultaneously.

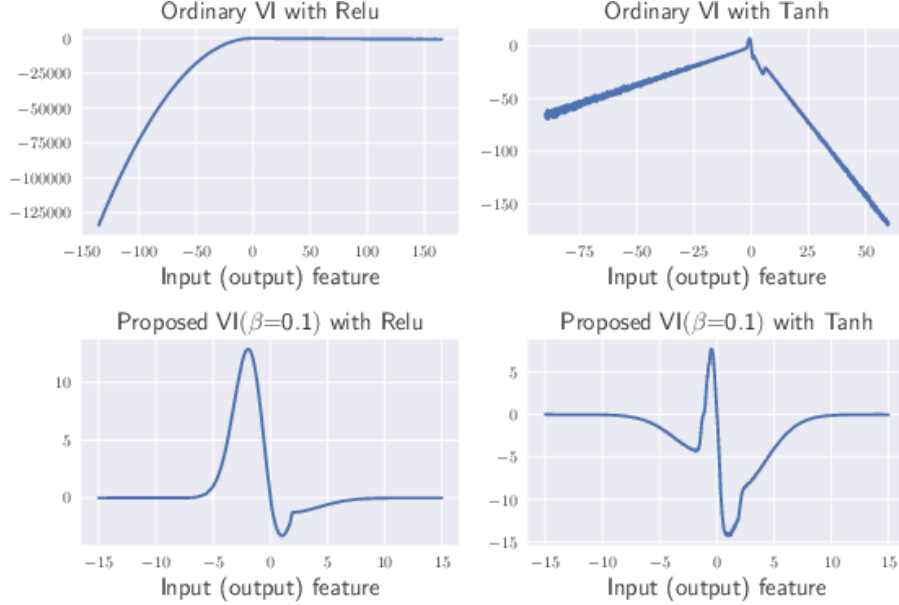

Figure 7: Perturbation on test log-likelihood for neural net regression.

From this figure, we confirmed again that in this situation, the perturbation on ordinary VI is not unbounded and the perturbation on our proposed method is bounded.

**Classification**

In the classification problem, first we considered how predictive distribution depends on the input related outlier. The method is as same as the regression problem. Since inputs have 14 dimensional features, $x \in \mathbb{R}^{14}$, we chose the third feature $x_3$ to move.

In the label misspecification experiment, we flipped one label of training dataset and measured how log-likelihood of test data are changed. From this experiment, we measured how label misspecification by chosen training data influences the prediction. We repeated this procedure for every training data point and took average. By this experiment, we measured how one flip of training data would influence the prediction on average.

The results shown in the main paper indicates that ordinary VI causes larger minus test log-likelihood change compared to $\beta$ VI. Base on the fact that decrease of log likelihood is almost equivalent to the increase of loss, the label misspecification causes larger perturbation to prediction in ordinary VI compared to proposed VI.

**Calculation of the Hessian**

In the above calculation, we have to evaluate the Hessian of ELBO. To save the computational cost we used following method,

$$\frac{\partial^2 L_\beta}{\partial m^2} v = \arg\min_{t} \frac{1}{2} t^\top \frac{\partial^2 L_\beta}{\partial m^2} t - v^\top t \quad (102)$$

This is the technique that instead of calculating the Hessian directly, we can calculate the product of the Hessian and a vector by solving the second order optimization problem. In our case, we consider $t = \frac{\partial}{\partial m} \mathbb{E}_{q^*(\theta)}\left[\ln p(x_{\text{test}}|\theta)\right]$ and solve above optimization problem.

**Influence function of the parameter of the neural network**

In this section, we show the IF of parameters. Figure 8 shows the plot of $IF(x_1, W, G)$ where W is a chosen one affine parameter in the case of *relu* activation function. Figure 8(a) shows the case of ordinary VI, which diverges as absolute value of $x_1$ become large. This means outliers have unlimited influence to the estimated static. On the other hand, Fig 8(b) shows the case of proposed method and the influence is bounded, that is the effect of outliers goes to zero. These results are compatible our theoretical analysis in the previous section.

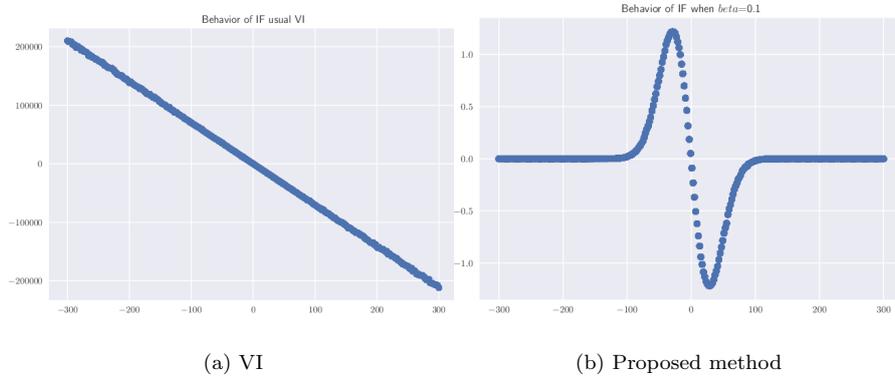

(a) VI      (b) Proposed method

Figure 8: IF of one affine parameter in Bayesian neural net.

However this is not sufficient analysis because what we want to be robust is the predictive distribution not parameters. Accordingly, it is necessary to study whether prediction is robust against outliers. For the analysis of prediction, we simulated the test log-likelihood. Actually, if the test log-likelihood has affected so much by an outlier, that is prediction on the test point is affected so much. Accordingly, such a model is not robust even under contamination of one outlier.

### L.3 Bench mark dataset

In this experiment, we determined $\beta$ and $\gamma$ by cross validation. For both regression and classification settings, the range of $\beta$ and $\gamma$ are from 0.1 to 0.9.

For WL(weighted likelihood proposed in Wang et al. [2017]), we considered Beta distribution for the prior of the weights and we used the method of ADVI for the optimization. For Rényi VI, we chosen $\alpha$ from the set of $\{-1.5, -1.0, -0.5, 0.5, 1.0, 1.5\}$ by cross-validation. For BB-$\alpha$, we chosen $\alpha$ from the set of $\{0, 0.25, 0.5, 0.75, 1.0\}$ by cross-validation. For Student-t distribution, we chose the degree of freedom from 3 to 10 by cross-validation.

In both of the regression and classification problem, we artificially increased the percentage of both input and output related outliers in the training dataset.

To make the input related outliers, we first specified which features of the input we would contaminate. In this experiment, for regression tasks, since input dimension is not so large, we contaminated all the input features. For classification tasks, if the training data has $D$ dimensional features, we randomly chose $D/2$ dimensions to contaminate. Next we randomly chose the data points we contaminate from training dataset. We contaminated the features by adding the Gaussian noise. Since the input data had been preprocessed by standardization, the noise we use is the Gaussian distribution which follows $\epsilon \sim N(0, 6)$. From the numerical calculation of influence function, we considered that the noise which has the value of "6" sigma variance can be regarded as an outlier.

For output related outlier, in the same way as the input related outlier, we randomly chosen the point which we would contaminate and add the Gaussian noise which follows $\epsilon \sim N(0, 6)$.

We optimized by using Adam and reparameterization trick. The learning rate of Adam was set to 0.01 and MC samples was 5 except for covertype dataset. For the covertype dataset, the learning rate of Adam was set to 0.001 and we used 20 MC samples. The minibatch size was set to 128.